\documentclass[11pt]{article}

\usepackage[final]{acl}

\usepackage{latexsym}
\usepackage{amsmath}
\usepackage{amsfonts}
\usepackage[most]{tcolorbox}
\usepackage{booktabs}
\usepackage{multirow}
\usepackage{todonotes}
\usepackage{enumitem}
\usepackage{placeins} 
\usepackage{needspace}

\usepackage{microtype}

\usepackage{inconsolata}

\usepackage{graphicx}
\graphicspath{{./}{latex/}}
\usepackage{subcaption}
\usepackage{caption}
\usepackage{fontspec}
\usepackage[bidi=default]{babel}
\babelprovide[import, main]{english}
\babelprovide[import]{arabic}
\IfFontExistsTF{[Amiri-Regular.ttf]}{%
  \babelfont[arabic]{rm}[
    Extension      = .ttf,
    UprightFont    = *-Regular,
    BoldFont       = *-Bold,
    ItalicFont     = *-Italic,
    BoldItalicFont = *-BoldItalic,
  ]{Amiri}%
}{%
  \IfFontExistsTF{[ScheherazadeNew-Regular.ttf]}{%
    \babelfont[arabic]{rm}[Extension=.ttf, UprightFont=*-Regular, BoldFont=*-Bold]{ScheherazadeNew}%
  }{%
    \PackageWarning{acl_latex}{No Arabic font found; Arabic text will use the main font}%
  }%
}
\newcommand{\ar}[1]{\foreignlanguage{arabic}{#1}}
\title{Can Dialects Be Steered Like Languages? \\  Sparse Neurons and Distributed Directions in Arabic LLMs}

\author{
  \textbf{Kareem Elozeiri\thanks{\ \ Equal contribution. Authors listed in alphabetical order of first name.}\textsuperscript{1}},
  \textbf{Mervat Abassy\footnotemark[1]\textsuperscript{1}},
  \textbf{Omar Kallas\textsuperscript{1}},
  \textbf{Fahim Dalvi\textsuperscript{2}},
\\
  \textbf{Preslav Nakov\textsuperscript{1}},
  \textbf{Kentaro Inui\textsuperscript{1,3,4}},
  \textbf{Nadir Durrani\textsuperscript{2}}
\\
\\
  \textsuperscript{1}Mohamed bin Zayed University of Artificial Intelligence
\\
  \textsuperscript{2}Qatar Computing Research Institute,
  Hamad Bin Khalifa University
\\
  \textsuperscript{3}Tohoku University
  \textsuperscript{4}RIKEN
\\
  \texttt{\{kareem.ali, mervat.abassy\}@mbzuai.ac.ae}
}

\newcommand{\allammodel}{\shortstack[l]{ALLaM-7B-\\Instruct}}
\newcommand{\fanarmodel}{\shortstack[l]{Fanar-1-9B-\\Instruct}}

\begin{document}
\maketitle
\begin{abstract}
Dialectal data are scarce relative to Modern Standard Arabic (MSA), causing Arabic LLMs to overproduce MSA and struggle with dialectally accurate generation. This raises a fundamental interpretability question about where and how dialectal features are encoded within model internals and whether these representations can improve dialect generation without fine-tuning. We study two inference-time approaches as interpretability probes and control mechanisms. First, neuron-level analysis identifies sparse populations that encode dialect-specific features and tests whether amplifying or suppressing them steers model outputs toward target dialects. Second, vector steering extracts dialect-specific activation directions and injects them during inference, motivated by feature entanglement at the neuron level. We find that these neurons are real but only partially explanatory. They occupy under 1\% of MLP dimensions but span only 5\% to 21\% of the residual dialect direction. This limited coverage is causally consequential. Neuron steering reinforces dialect in some varieties when the prompt is already dialectal but cannot induce it from MSA prompts, whereas vector steering succeeds in both settings. Arabic dialects are therefore steerable mainly through distributed rather than localized representations.\footnote{Code and artifacts are available at \url{https://github.com/mbzuai-nlp/arabic-dialect-steering}}
\end{abstract}

\section{Introduction}
\label{sec:intro}

Arabic large language models (LLMs) have made strong progress on standard benchmarks, yet remain weak on Arabic dialects \cite{mousi-etal-2025-aradice,robinson-etal-2025-al}. This is partly reflected by the pretraining data where Modern Standard Arabic (MSA) dominates most Arabic corpora, while dialectal text is scarce and inconsistently sourced \cite{bari2024allamlargelanguagemodels}. 

As a result, models often default to MSA or produce hybrid outputs that indiscriminately blend dialects. \citet{nacar2025} documents this pattern in ALLaM 34B, where Hijazi and Egyptian prompts consistently elicit MSA-style responses, limiting the model’s usefulness for conversational agents, cultural content generation, and other applications requiring dialectal authenticity.


\begin{figure*}[t]
    \centering
    \includegraphics[width=\textwidth]{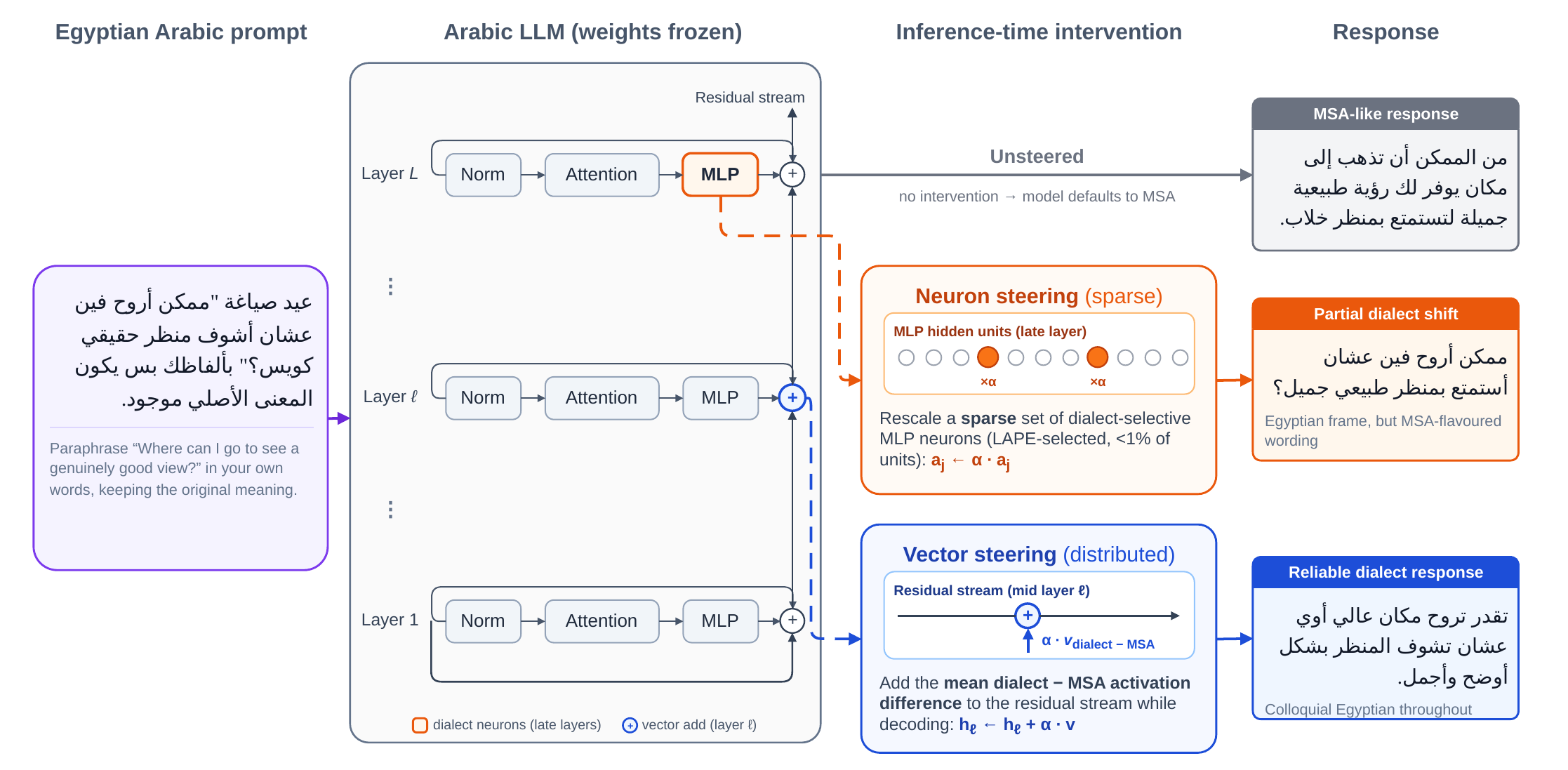}
    \caption{Two inference-time interventions for Arabic dialect control on a frozen model. Neuron steering rescales a sparse set of dialect-selective MLP units; vector steering adds a dialect--MSA activation difference to the residual stream. Example shows responses go from MSA (unsteered) to a partial shift (neuron) to consistent Egyptian (vector).}
    \label{fig:overview}
\end{figure*}

Recent work on interpretability suggests that linguistic behavior in LLMs can be localized or linearly controlled: neurons can encode language-relevant features, and activation-space directions can steer generation toward target attributes \cite{JMLR:v24:23-0074,tang-etal-2024-language,rahmanisa2025language_specific,turner2024activation,panickssery2024steering,chen2025persona}. These findings raise a question for Arabic dialects: \textit{Do dialectal features admit similar localized control, or does the lexical and syntactic continuity among Arabic varieties make them more distributed?} This distinction matters because dialects are not separate languages with clean boundaries; they share vocabulary, morphology, and orthography with MSA and with each other.

This task presents unique challenges: unlike independent languages, Arabic dialects have high lexical and syntactic overlap, and dialectal boundaries are fuzzy rather than discrete \cite{habash2010introduction}, making it harder to disentangle dialectal features at the neuron level \cite{abdelali-etal-2022-post}. To address these open challenges, this work investigates two distinct yet complementary inference-time strategies for dialect control in Arabic LLMs, illustrated in Figure~\ref{fig:overview}.. The first investigates whether Arabic LLMs encode dialect-specific neurons and whether manipulating these neurons can steer generation toward the desired dialect. The second takes a broader representational perspective, asking whether dialect-specific directions can be extracted from the model's activation space and injected during inference to achieve more robust dialectal fidelity without fine-tuning or modification.





\paragraph{Research Questions:} We ask three questions:
\begin{enumerate}[leftmargin=*]
    \item Can Arabic dialects be steered in the same way as languages, despite their lexical and syntactic overlap with MSA and with each other?
    \item Are dialectal features localized in neuron populations, distributed across residual space, or both?
    \item Which inference-time intervention provides more reliable dialect control: neuron-level steering or activation-vector steering?
\end{enumerate}

Our results reveal a nuanced picture: Arabic dialects are neither encoded as fully isolated neuron-level modules nor as entirely diffuse behaviors. Instead, dialectal information is both sparse and distributed. We find that dialect-associated neurons concentrate in late generation-facing layers, with MSA clearly separated from spoken dialects and with substantial sharing among regional dialects. However, these sparse neurons capture only a partial projection of the broader residual-space dialect direction. This explains why neuron steering can reinforce dialectal behavior, but distributed activation-vector steering provides more reliable control. Together, these findings suggest that Arabic dialect variation forms a structured and causally steerable dimension of LLM representation space. To the best of our knowledge, this is the first study to show that Arabic dialects are causally steerable within LLMs through both sparse neuron-level mechanisms and distributed activation-space directions. 



\section{Methodology}

We study two inference-time interventions for Arabic dialect control. Neuron-based steering operates on sparse dialect-associated MLP neurons, while vector steering operates on dialect-specific directions extracted from contrastive dialect--MSA pairs. Both keep model weights fixed and serve as causal probes of dialectal representations. We also measure their inference-time latency overhead
(Appendix~\ref{app:runtime-overhead}).

\subsection{Neuron-Based Dialect Steering}
\label{sec:neuron-steer}


We test whether dialectal behavior can be controlled through sparse neuron-level interventions. Following the LAPE-based framework \cite{tang-etal-2024-language}, we identify dialect-associated MLP neurons and rescale their activations during decoding to steer generation toward target dialects.

\paragraph{Neuron Identification:}


We treat dialect-associated neurons as those that are frequently active for one Arabic dialect but selective across dialects. Let $L$ be the set of dialects used for extraction, $S_k$ the tokenized corpus for dialect $k\in L$, and $T_k=\{(s,t):s\in S_k,\;1\leq t\leq |s|\}$ the set of sentence--token-position pairs in that corpus. For neuron $j$ in layer $i$, with scalar activation $\operatorname{act}(i,j;s,t)$, we estimate its activation probability for dialect $k$ as $\hat{p}^{k}_{i,j}=\mathbb{E}_{(s,t)\in T_k}\left[\mathbf{1}\{\operatorname{act}(i,j;s,t)>0\}\right]$. 

We collect these probabilities across dialects as $p_{i,j}=[\hat{p}^{k}_{i,j}]_{k\in L}$ and normalize the vector to unit $\ell_1$ norm, yielding $p'_{i,j}$. The LAPE score is the entropy of this normalized distribution: $\operatorname{LAPE}_{i,j}=-\sum_{k\in L}{p'}^{k}_{i,j}\log {p'}^{k}_{i,j}$. Low LAPE indicates that a neuron is selective for a small number of dialects. We select a neuron as dialect-associated if it falls in the top $n\%$ of activation probability for at least one dialect and in the bottom $m\%$ of LAPE scores across neurons, where $n$ and $m$ are selection-percentile hyperparameters. These neurons are then used as targets for inference-time steering.

\paragraph{Neuron Steering:}

We steer the model at inference time by rescaling the activations of selected dialect-associated MLP neurons during decoding, leaving model parameters unchanged. For a target dialect $k$, let $\mathcal{D}_k(i)$ denote the set of selected target-dialect neurons in layer $i$. We define $\mathcal{D}_{\mathrm{MSA}}(i)$ as the selected MSA neuron set and, optionally, $\mathcal{D}_{\mathcal{C}}(i)$ as the union of selected neurons for non-target competitor dialects. Let $a_{i,t,j}$ be the activation of neuron $j$ in layer $i$ at decoding step $t$. We apply the following multiplicative update:
\[
\tilde{a}_{i,t,j}
=
\lambda_{i,j} a_{i,t,j},
\]
where
\[
\lambda_{i,j}
=
\begin{cases}
\alpha, & j \in \mathcal{D}_k(i),\\
\gamma, & j \in \mathcal{D}_{\mathrm{MSA}}(i),\\
\gamma_{\mathrm{comp}}, & j \in \mathcal{D}_{\mathcal{C}}(i),\\
1, & \text{otherwise.}
\end{cases}
\]

Here, $\alpha$ is the target-dialect amplification factor,
$\gamma$ is the MSA suppression factor, and $\gamma_{\mathrm{comp}}$ is the optional competitor-dialect suppression factor. We use $\alpha>1$ to strengthen target-dialect neurons and $\gamma,\gamma_{\mathrm{comp}}<1$ to suppress MSA and non-target dialect neurons. Thus, decoding is biased toward the target dialect while weights remain unchanged.

\subsection{Vector Steering}
\label{sec:vector-steer}
While neuron-level steering offers fine-grained 
interpretability, dialect features in Arabic LLMs may be distributed across many neurons 
simultaneously, making isolated neuron interventions insufficient. To address this, we 
explore a complementary representation-level approach: \textit{vector steering} \cite{turner2024activation, panickssery2024steering}, which 
operates on the full activation space of a layer rather than on individual neurons.

\paragraph{Steering Vector Extraction:}
For a target dialect $k$ and a contrastive variety (MSA), we collect a parallel corpus 
of sentence pairs $\{(s^k_i, s^{\text{MSA}}_i)\}_{i=1}^{N}$. Each sentence is passed through the model independently, and we extract the mean hidden-state activation across all response tokens at layer $\ell$, denoted $h^\ell(s)$. The dialect steering vector is then computed as the mean 
difference between dialect and MSA activations:

\vspace{-2mm}

\begin{equation}
    v^k_\ell = \frac{1}{N} \sum_{i=1}^{N} 
    \left( h^\ell(s^k_i) - h^\ell(s^{\text{MSA}}_i) \right)
    \label{eq:steering_vector}
    \tag{Equation 1}
    \nonumber
\end{equation}

This difference vector encodes the directional shift in activation space corresponding 
to moving from MSA toward dialect $k$, capturing 
distributed dialect features that may not be localized to any individual neuron. 
Crucially, because activations are extracted from the model's response tokens rather 
than the input token, the vector reflects the model's \textit{generated} dialectal 
behavior rather than its encoding of the input, making it a more direct target for 
generation-time intervention.


\paragraph{Inference-Time Injection:} During generation, we register a  forward hook at layer $\ell$ that intercepts the activation tensor at every token generation step and adds the scaled steering vector in-place, $h^\ell_{\text{new}} = h^\ell + \alpha \cdot v^k_\ell$, where $\alpha$ is a scalar coefficient controlling the strength of the intervention. Downstream layers then receive the modified activations and generate dialect-influenced tokens accordingly. No retraining or parameter updates are required.





\section{Experiments}
\label{sec:experiments}

\subsection{Experimental Setup}

\paragraph{Models:}
We experiment with two Arabic-centric large language models:
ALLaM-7B-Instruct-preview \cite{bari2024allamlargelanguagemodels} and
Fanar-1-9B-Instruct \cite{fanarllm2025}. Both are instruction-tuned specifically for
Arabic. ALLaM comprises 32 transformer layers, while Fanar has 42.
Experimenting across both models allows us to assess whether findings
generalize across architectures trained on different Arabic corpora and
with different instruction-tuning procedures. 

\paragraph{Dialects and data:}
Ablation experiments are conducted on Egyptian Arabic (Cairo) and Moroccan Arabic (Rabat) as these dialects offer both larger amounts of parallel data.

Parallel data is essential for steering vector extraction and direct evaluation availability in the benchmark used in this work. Additionally, they represent linguistically diverse dialects with substantial variation between them, allowing for a more informative analysis of the proposed methods.

Final model performance is then evaluated across four broader dialect groups: Egyptian, Moroccan, Levantine, and Gulf Arabic. Steering vectors and dialect-specific neurons are extracted using parallel dialect data from the MADAR corpus. Specifically, Cairo, Rabat, Beirut, and Doha each provide 12,000 sentence pairs, which are used for steering vector extraction and neuron identification.
In contrast, Riyadh and Aleppo provide only 2,000 sentence pairs each. For evaluation, Cairo and Rabat are mapped directly to the Egyptian and Moroccan benchmark subsets. Beirut and Aleppo are evaluated using the Syrian subset, as they are the closest available dialects in the benchmark, while Doha and Riyadh are evaluated on the Saudi subset.

\paragraph{Evaluation:}
\label{sec:eval-setup}
Primary evaluation uses the LLM-as-a-judge protocol described in
Section~\ref{sec:llm-judge}. We additionally evaluate using the AL-QASIDA
benchmark as a complementary automatic metric. For the best-performing model
configurations for each dialect, human annotations are collected to measure inter-rater
agreement with the LLM judge.

\subsection{LLM-as-a-Judge}
\label{sec:llm-judge}
We assess
generation quality, using Gemini 2.5 Flash
\cite{comanici2025gemini25pushingfrontier} as the judge. Each response is
evaluated independently against the target dialect across four integer-scored
(1--5) dimensions: \textit{dialect authenticity}, \textit{coherence},
\textit{Arabic fluency}, and \textit{MSA formality}. The first three serve as
quality measures while MSA formality is used diagnostically to capture drift toward
formal Modern Standard Arabic. We report mean scores per dimension over all judged samples. The
full prompt, rubric, and implementation details are in
Appendix~\ref{app:llm-judge-setup}.

\subsection{Human Evaluation}
\label{sec:human_eval}

We recruited eight Arabic annotators in total, assigning two annotators to each of the four target dialects. The evaluation covered 300 outputs for every combination of dialect, model, and generation condition (unsteered, neuron steering, and vector steering), yielding \(300 \times 4 \times 2 \times 3 = 7{,}200\) evaluated outputs. Each output was scored independently by both annotators assigned to its target dialect.

The same four 1--5 dimensions as the LLM-as-a-judge evaluation were used and model identities were hidden from the annotators to reduce bias. For each output and metric, we first average the two annotator scores and then average these item-level scores across all samples. We report the human average as the mean of Arabic fluency, coherence, and dialect authenticity, while reporting MSA formality separately.

\subsection{AL-QASIDA Evaluation}
\label{sec:alqasida}

We evaluate using the AL-QASIDA framework~\cite{robinson-etal-2025-al}, scoring
responses with the ADI2 metric:
\begin{align*}
\text{ADI2}(y) &= P(y \text{ is dialectal}) \times P(y \text{ is dialect } C)
\end{align*}
which jointly penalizes MSA responses and responses in the wrong dialect. We
also report macro-ADI2, which aggregates dialect probabilities at the regional
level to account for overlap between geographically proximate varieties. We use
the monolingual generation task of the benchmark throughout, as it best reflects real-world
usage and aligns with our inference-time intervention. Full details on the
framework, prompt construction, and corpora are in
Appendix~\ref{app:evalalqas}.
\subsection{Neuron-based Steering}

\paragraph{Neuron identification:}
Following the selection procedure defined in
Section~\ref{sec:neuron-steer}, we select neurons that fall in the top 5\% of activation
probability for at least one dialect and in the bottom 1\% of LAPE-score distribution
across all neurons. These thresholds keep neurons that are both frequently active for a dialect and highly selective across dialects.

\begin{figure}[t]
    \centering

    \begin{subfigure}{\columnwidth}
        \centering
        \includegraphics[width=\columnwidth]{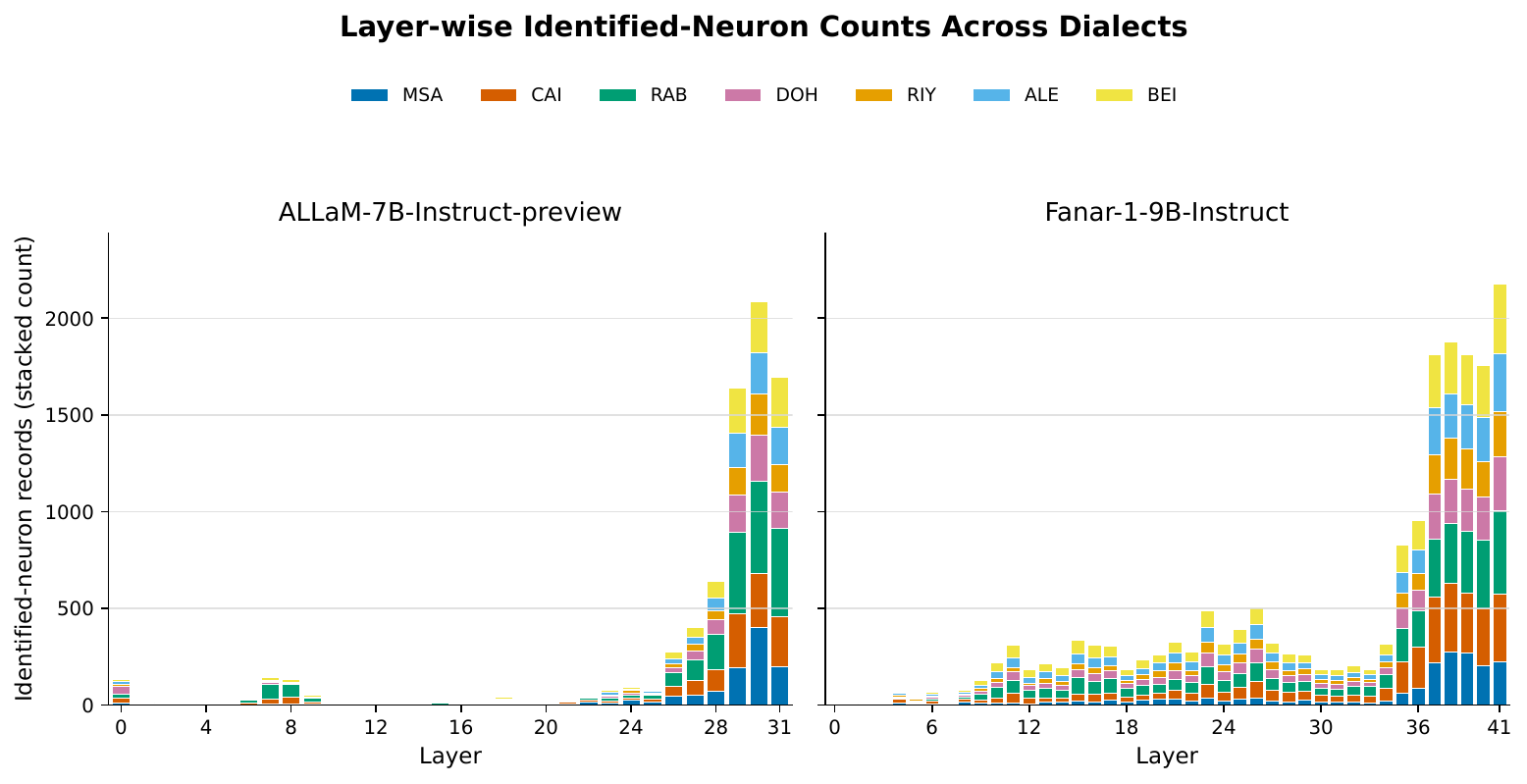}
        \caption{}
        \label{fig:identified-neurons-layer}
    \end{subfigure}

    \vspace{0.4em}

    \begin{subfigure}{\columnwidth}
        \centering
        \includegraphics[width=\columnwidth]{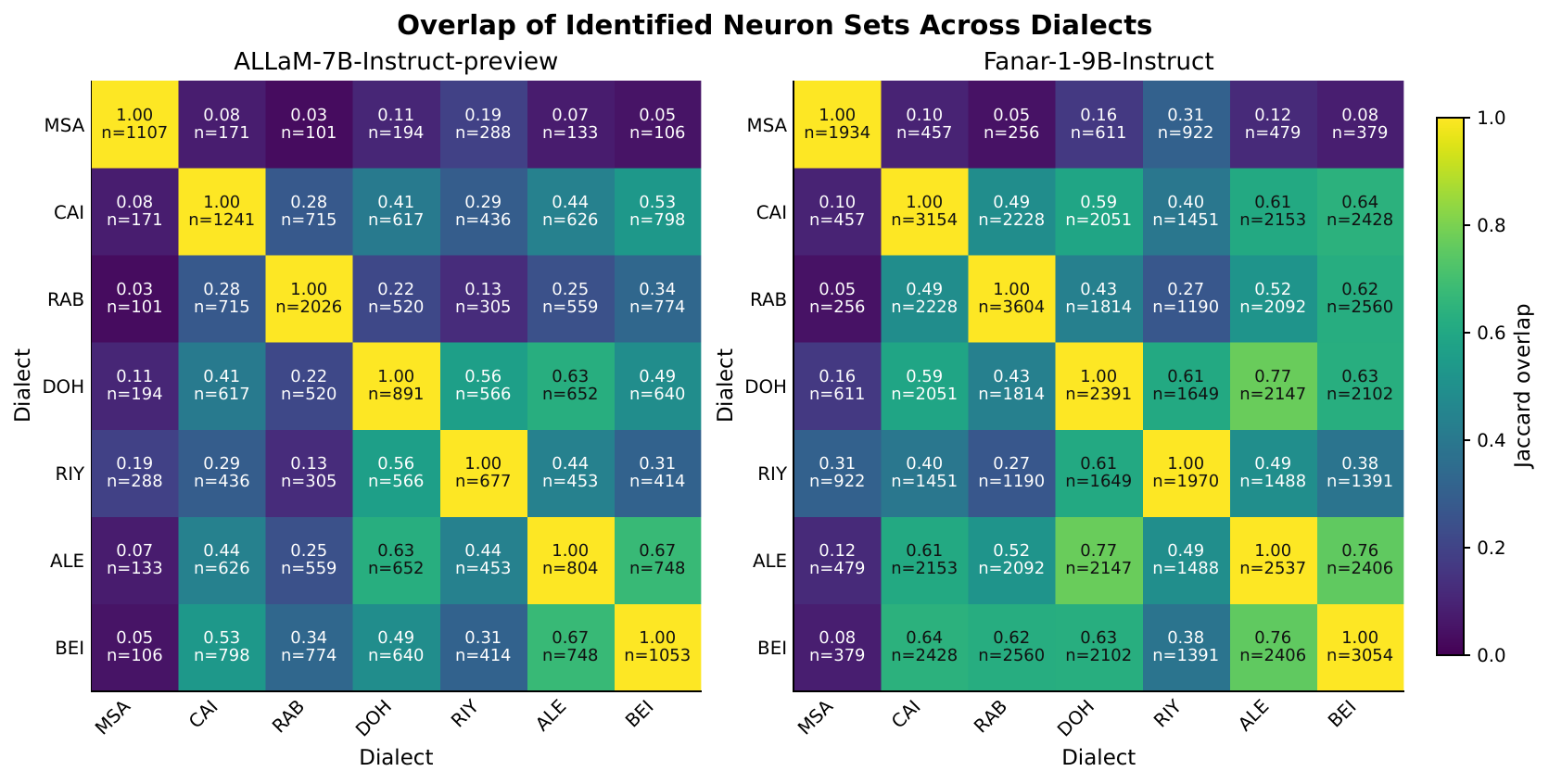}
        \caption{}
        \label{fig:identified-neurons-overlap}
    \end{subfigure}

   \caption{
Dialect-specific neuron analysis for ALLaM and Fanar across MSA, Cairo, Rabat,
Doha, Riyadh, Aleppo, and Beirut. Top: identified neurons concentrate in later
layers. Bottom: Jaccard overlap shows stronger sharing among spoken dialects
than with MSA.
}
    \label{fig:identified-neurons-analysis}
\end{figure}

\paragraph{Layer-wise distribution:}
Figure~\ref{fig:identified-neurons-layer} shows that dialect-specific neurons
are depth-localized rather than uniformly distributed. In ALLaM, they form a
narrow late-layer bottleneck, with layers 29--31 accounting for 69.5\% of all
dialect-neuron records and layer 30 as the main peak. Fanar shows a similar
late-layer pattern, but spread over a wider terminal band, with layers 37--41
accounting for 50.6\% of records, plus a visible mid-layer tail. This suggests
that dialectal behavior is not primarily encoded as early surface cues such as
orthography or local lexical features. Instead, it emerges closer to the
generation-facing layers where internal Arabic representations are mapped into
next-token choices, making these neurons plausible control points for dialectal
style, lexical choice, and morphosyntactic realization.


\paragraph{Cross-dialect overlap:}
Figure~\ref{fig:identified-neurons-overlap} shows that the identified neuron
sets are neither isolated dialect modules nor a single generic Arabic circuit.
MSA is consistently separated from the dialects, with average MSA--dialect
overlap of only 0.09 in ALLaM and 0.14 in Fanar, compared with dialect--dialect
overlap of 0.40 and 0.55. Since all varieties share substantial Arabic lexical
and orthographic structure, this separation suggests that the selected neurons
are not merely generic Arabic detectors. Instead, dialects share part of their
internal representation while retaining distinct neuron sets. High-overlap pairs
such as Aleppine--Beiruti and Doha--Riyadh suggest reusable regional
subcircuits that may support transfer, while low overlap with MSA marks sharper
boundaries where MSA-based interventions are less likely to transfer.

\paragraph{Ablations:}
We ablate the target-dialect amplification factor $\alpha$, the MSA
suppression factor $\gamma$, and the competitor-dialect suppression factor
$\gamma_{\mathrm{comp}}$, varying one parameter while keeping the other two
fixed. Ablations are run on Egyptian and Moroccan Arabic across both models.
Our results show that $\alpha$ is the main parameter affecting dialectal
generation performance, with the best setting being $\alpha = 2.0$ for
ALLaM and $\alpha = 4.0$ for Fanar.

Varying $\gamma$ and
$\gamma_{\mathrm{comp}}$ does not yield consistent improvements and in some
cases degrades performance. Full ablation settings and results are reported
in Appendix~\ref{app:neuron-steer-abls}.

\subsection{Vector Steering}

To construct the parallel sentence pairs $\{(s^k_i, s^{\text{MSA}}_i)\}$
required by ~\ref{eq:steering_vector}, we use the MADAR corpus. Each
dialect sentence is placed in the assistant turn of the chat template,
formatted as if produced by the model, so that hidden states are extracted
from response token positions, which is consistent with how vectors are applied at
generation time.

Before applying steering vectors at inference time, we first ask a more fundamental interpretability question: \emph{do the hidden states of Arabic LLMs encode dialect identity in a structured and geometrically separable way?} If dialect-specific directions exist in activation space, they should be recoverable from the mean hidden-state representations used to construct our steering vectors. We investigate this through PCA of the extracted representations at a fixed intermediate layer across three Arabic LLMs (full analysis in Appendix~\ref{app:vector_analysis}). We find that dialects form geographically coherent clusters in representation space, with dialect varieties occupying distinct regions, and that representational distance roughly tracks real-world linguistic proximity. This structured geometry confirms that dialect identity is encoded in a recoverable and separable way, motivating the use of mean activation differences as steering directions.

\paragraph{Ablations:}
We conduct ablations across layers, coefficients and number of steered tokens.

\begin{figure}[t]
    \centering
    \begin{subfigure}{\columnwidth}
        \centering
        \includegraphics[width=\columnwidth]{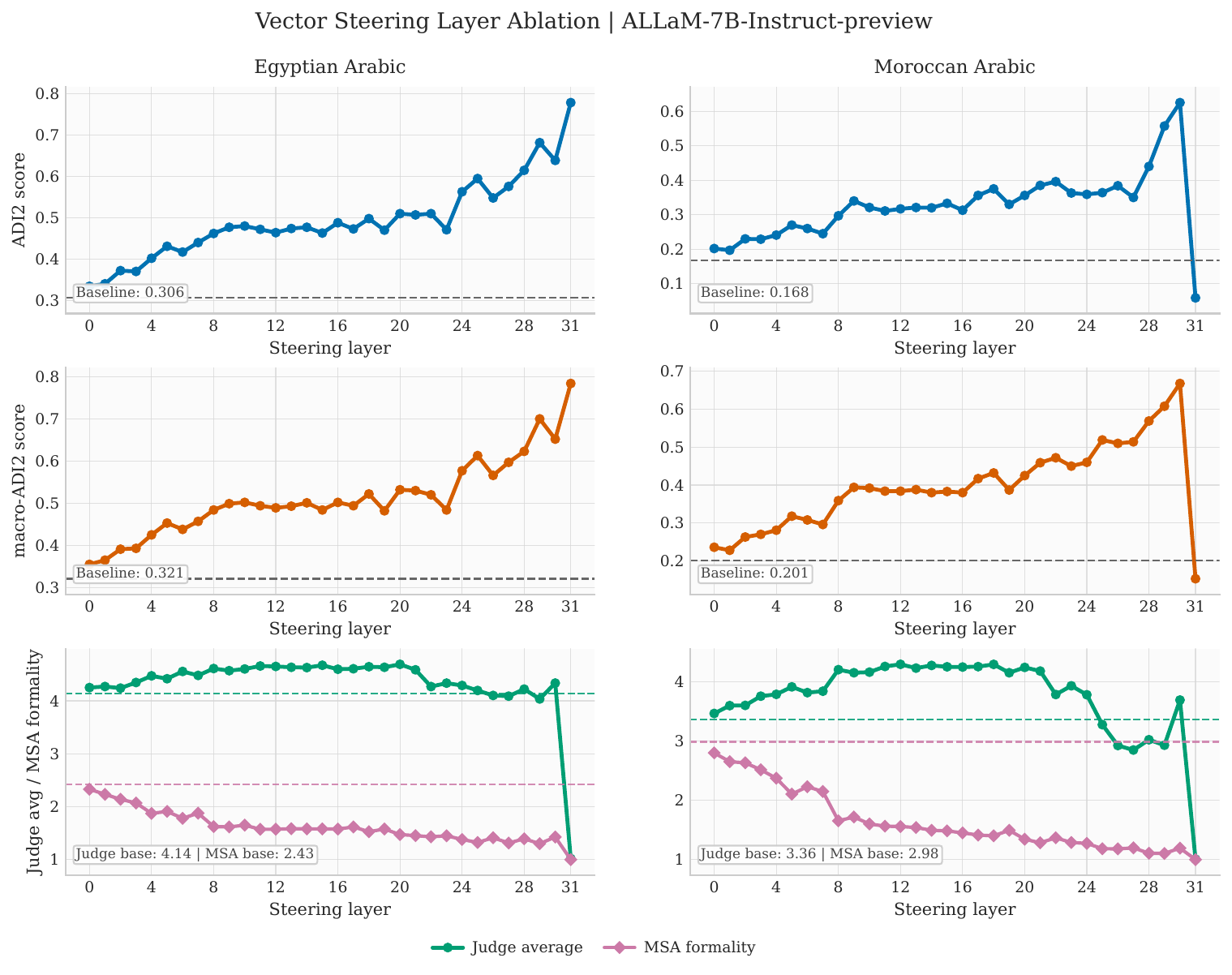}
        \caption{ALLaM-7B-Instruct-preview}
        \label{fig:layer_ablation_allam}
    \end{subfigure}
    \vspace{0.5em}
    \begin{subfigure}{\columnwidth}
        \centering
        \includegraphics[width=\columnwidth]{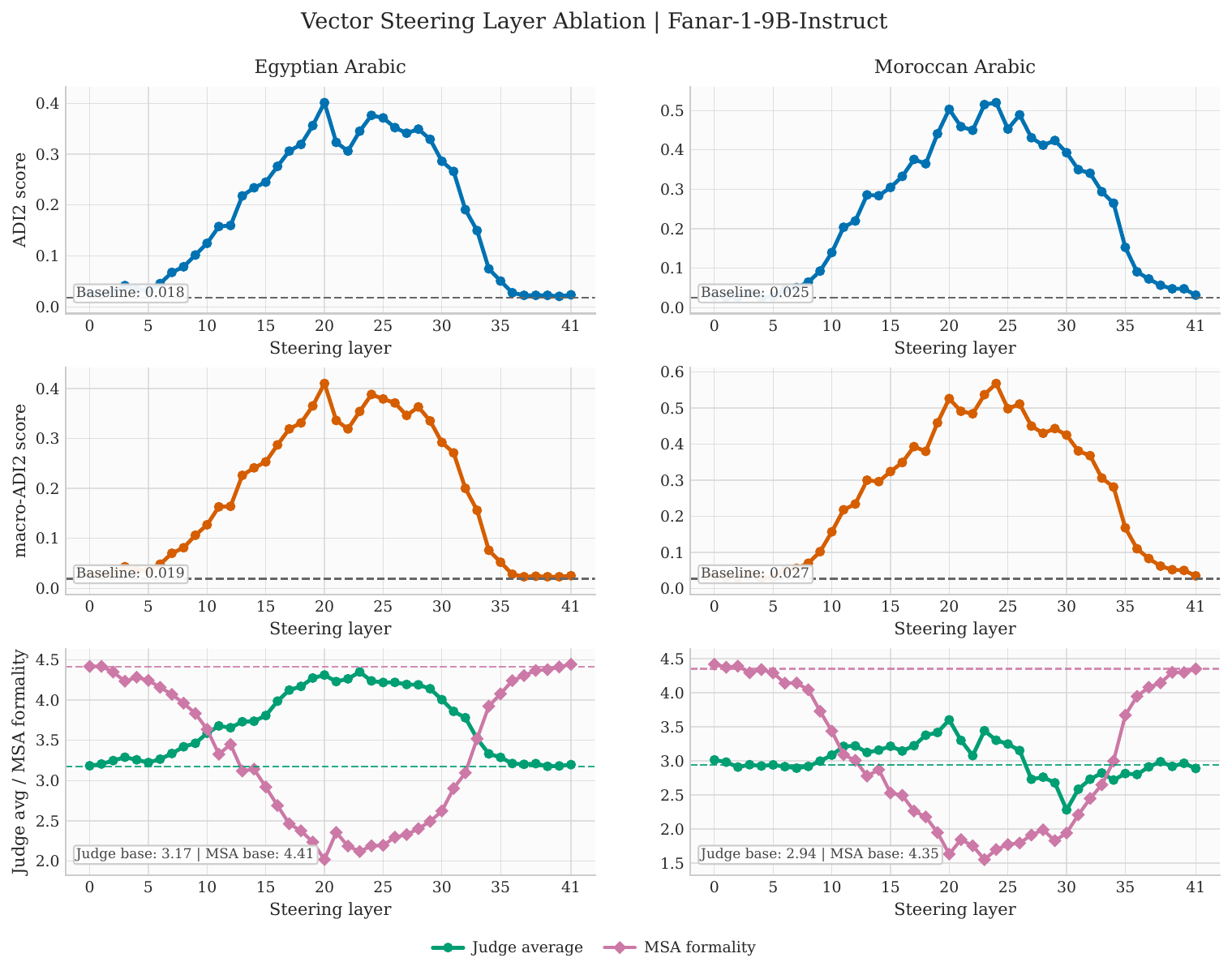}
        \caption{Fanar-1-9B-Instruct}
        \label{fig:layer_ablation_fanar}
    \end{subfigure}
    \caption{Layer ablation on Egyptian and Moroccan Arabic with a fixed
    coefficient $\alpha = 2$. ADI2, macro-ADI2, LLM judge average, and MSA
    formality are shown across all layers for each model. Dashed lines
    indicate unsteered baselines.}
    \label{fig:layer_ablation}
\end{figure}

\textit{Layer selection:}
All layer ablations are conducted with a fixed steering coefficient of
$\alpha = 2$. As shown in Figure~\ref{fig:layer_ablation}, the two models
reveal strikingly different dialectal encoding profiles. Fanar concentrates
dialect information in its middle layers, with both ADI2 and macro-ADI2
peaking around layers 19--24; the LLM judge corroborates this, showing a
corresponding rise in dialect quality scores that peaks in the same region
before declining. ALLaM tells a different story: judge scores remain broadly
stable across most layers, suggesting that generation quality is largely
insensitive to intervention depth, while ADI2 rises monotonically, most
sharply at the final layers, indicating that benchmark gains at depth come
without a corresponding perceptual improvement.

In both models, MSA formality
moves inversely with dialect quality scores, confirming that the steering
vector shifts register rather than introducing unrelated artifacts. Based on
the highest LLM-as-a-judge score, we select candidate layers from the high-quality plateau of
each model (layers 19--24 for Fanar, 11--20 for ALLaM) for the subsequent
coefficient ablation.

\textit{Coefficient selection:}
$\alpha$ is swept over ${1, 2, 3, 4, 5}$ across the candidate layers (Figure~\ref{fig:coeff_ablation}, Appendix~\ref{app:coeff_ablation}). ALLaM and Fanar differ in robustness: Fanar's Egyptian Arabic shows no divergence between ADI2 and judge scores even at $\alpha = 5$, whereas ALLaM degrades in coherence at higher coefficients. Moroccan Arabic is more sensitive than Egyptian Arabic for both models, with judge quality declining while ADI2 continues to rise. The best layer--coefficient pair per model per dialect is carried forward to all subsequent experiments.

\textit{Token budget:}
Using the optimal layer and coefficient per model--dialect pair, steering is restricted to the first $N$ decoded tokens, with $N$ increased in steps of 10 (Appendix~\ref{app:token_budget}). The two models behave differently: ALLaM commits to the target dialect after minimal intervention, with judge scores and ADI2 plateauing from as few as 10--20 tokens onward, suggesting the model sustains the dialect autonomously once primed. Fanar requires sustained injection, with ADI2 and macro scores rising steadily with $N$, consistent with its near-zero unsteered baseline. Full-sequence steering on Moroccan Arabic for Fanar reveals a tension between dialect authenticity and output quality that ADI2 alone would not expose.

\textit{Extraction-size stability:} We test whether steering directions are
stable across extraction-corpus sizes by re-estimating vectors from 1k--12k
parallel pairs and comparing layer-wise cosine similarity. The directions
stabilize by 4k examples at the selected steering layers, suggesting that vector
orientation is robust to sampling variation; full results are in
Appendix~\ref{app:vect-steer-sensitivity}.
\section{Results}
\label{sec:results}

\begin{table*}[t]
\centering
\scriptsize
\setlength{\tabcolsep}{2pt}
\begin{tabular*}{\textwidth}{@{\extracolsep{\fill}}llcccccccccccc@{}}
\toprule
& & \multicolumn{4}{c}{ADI2} & \multicolumn{4}{c}{macro-ADI2} & \multicolumn{4}{c}{Judge avg.} \\
\cmidrule(lr){3-6}\cmidrule(lr){7-10}\cmidrule(lr){11-14}
Model & Method
& EGY & MOR & SAU & SYR
& EGY & MOR & SAU & SYR
& EGY & MOR & SAU & SYR \\
\midrule
\multirow{4}{*}{\shortstack[l]{ALLaM-7B-\\Instruct}}
& Unsteered
& 0.306 & 0.168 & 0.036 & 0.094
& 0.321 & 0.201 & 0.084 & 0.252
& 4.137 & 3.359 & 3.580 & 3.873 \\
& Explicit
& \textbf{0.631} & \textbf{0.518} & \textbf{0.194} & \textbf{0.260}
& \textbf{0.640} & \textbf{0.570} & \textbf{0.387} & \textbf{0.613}
& \underline{4.677} & \textbf{4.469} & \textbf{4.298} & \textbf{4.527} \\
& Neuron
& 0.468 & 0.308 & 0.027 & 0.062
& 0.471 & 0.318 & 0.072 & 0.228
& 3.978 & 3.766 & 3.282 & 3.494 \\
& Vector
& \underline{0.512} & \underline{0.310} & \underline{0.124} & \underline{0.156}
& \underline{0.532} & \underline{0.380} & \underline{0.252} & \underline{0.407}
& \textbf{4.712} & \underline{4.390}& \underline{4.151} & \underline{4.470} \\
\midrule
\multirow{4}{*}{\shortstack[l]{Fanar-1-9B-\\Instruct}}
& Unsteered
& 0.018 & 0.025 & 0.004 & 0.006
& 0.019 & 0.027 & 0.022 & 0.017
& 3.175 & 2.944 & \underline{3.462} & 3.172 \\
& Explicit
& \underline{0.383} & \underline{0.345} & \textbf{0.028} & \textbf{0.158}
& \underline{0.388} & \underline{0.347} & \textbf{0.076} & \textbf{0.279}
& \underline{3.986} & \underline{3.599} & 3.396 & \underline{3.509} \\
& Neuron
& 0.119 & 0.070 & 0.002 & 0.007
& 0.119 & 0.078 & 0.020 & 0.017
& 3.029 & 2.580 & 3.244 & 2.974 \\
& Vector
& \textbf{0.384} & \textbf{0.410} & \underline{0.010} & \underline{0.044}
& \textbf{0.398} & \textbf{0.428} & \underline{0.035} & \underline{0.200}
& \textbf{4.446} & \textbf{4.060} & \textbf{3.536} & \textbf{3.648} \\
\bottomrule
\end{tabular*}
\caption{
Mono-dialect results across automatic and LLM-as-a-judge metrics. Higher is
better for ADI2, macro-ADI2, and judge average. EGY, MOR, SAU, and SYR denote
Egyptian, Moroccan, Saudi, and Syrian Arabic, respectively. Neuron and Vector
denote the best neuron-steering and vector-steering configurations. Within each
model, metric, and dialect, the best score is shown in bold and the second-best
score is underlined.
}
\label{tab:mono_summary_scores}
\end{table*}

\begin{table*}[t]
\centering
\scriptsize
\setlength{\tabcolsep}{3pt}
\begin{tabular*}{\textwidth}{@{\extracolsep{\fill}}llcccccccc@{}}
\toprule
& & & & \multicolumn{3}{c}{Human avg. $\uparrow$} & \multicolumn{3}{c}{MSA formality $\downarrow$} \\
\cmidrule(lr){5-7}\cmidrule(lr){8-10}
Model & Dialect & Layer & $\alpha$ & Unst. & Neu. & Vec. & Unst. & Neu. & Vec. \\
\midrule
\multirow{4}{*}{ALLaM-7B-Instruct-preview}
& EGY & 20 & 2
& 3.698 & \underline{4.111} & \textbf{4.231}
& 3.195 & \underline{2.470} & \textbf{1.795} \\
& MOR & 19 & 1
& 3.300 & \underline{3.621} & \textbf{3.628}
& 4.165 & \underline{3.905} & \textbf{3.722} \\
& SAU & 20 & 2
& \underline{3.538} & 3.438 & \textbf{4.066}
& \underline{4.410} & 4.473 & \textbf{3.005} \\
& SYR & 20 & 2
& 3.693 & \underline{3.724} & \textbf{4.138}
& 3.177 & \underline{3.075} & \textbf{1.945} \\
\midrule
\multirow{4}{*}{Fanar-1-9B-Instruct}
& EGY & 23 & 3
& \underline{3.506} & 3.339 & \textbf{3.858}
& 4.953 & \underline{4.430} & \textbf{2.393} \\
& MOR & 20 & 2
& \textbf{3.005} & \underline{2.864} & 2.657
& 4.340 & \underline{3.998} & \textbf{2.707} \\
& SAU & 20 & 2
& \textbf{3.333} & 3.259 & \underline{3.285}
& \underline{4.723} & 4.842 & \textbf{4.620} \\
& SYR & 20 & 2
& \underline{3.526} & 3.221 & \textbf{3.617}
& 4.292 & \underline{4.210} & \textbf{2.825} \\
\bottomrule
\end{tabular*}
\caption{
Human evaluation summary for mono-dialect outputs. Each score is averaged over
two blinded annotators and then over samples. Human avg. averages fluency,
coherence, and dialect authenticity, while lower MSA formality is better. Layer
and $\alpha$ denote the vector-steering layer and coefficient. Bold and
underline mark the best and second-best values within each model and dialect.
Unst., Neu., and Vec. denote unsteered, neuron steering, and vector steering.
}
\label{tab:human_eval_mono_summary}
\end{table*}

We evaluate two generation settings. In the \emph{mono-dialect} setting, the
prompt is written in the target dialect, reflecting the  use case where a
user writes in a dialect and expects a response in kind. In the \emph{MSA-prompt}
setting, the prompt is written in MSA while the response is steered toward a
target dialect, testing whether internal steering can override the prompt
variety rather than simply preserve it.

\paragraph{Final evaluation setup:} All methods use deterministic decoding
with a maximum of 128 new tokens. The unsteered baseline receives the original
prompt without intervention. Neuron steering uses the LAPE-selected neurons with
$\alpha=2.0$ for ALLaM and $\alpha=4.0$ for Fanar, without MSA or competitor
suppression. Vector steering injects response-side dialect--MSA directions for
the first 30 generated tokens, following the token ablation results. This token
budget is stable for ALLaM and avoids the quality degradation observed for Fanar
under longer Moroccan steering. For Egyptian and Moroccan, we report the best
layer and coefficient settings from the ablations. For non-ablated dialects, we
use layer 20 as the default based on its strong performance across the ablated
settings. 

\subsection{Mono-Dialect Prompts}
\label{sec:results_mono_dialect}
Table~\ref{tab:mono_summary_scores} shows that vector steering demonstrates clear effectiveness, matching or surpassing explicit prompting on judge averages for Fanar across all dialects and for ALLaM on Egyptian, with minimal difference with explicit prompting on the remaining three dialects, despite explicit prompting (Details in Appendix~\ref{app:explicit-prompt}) leading on raw ADI2 and macro-ADI2. This dissociation reinforces that ADI2 is not always a representative metric, as it does not account for coherence or overall output quality. Neuron steering improves over the unsteered baseline, particularly for Egyptian and Moroccan for ADI2 scores but shows no improvement in LLM as a judge scores except for ALLaM on Moroccan. Human evaluation (Table~\ref{tab:human_eval_mono_summary}) validates these findings on authenticity and MSA formality in all eight groups, with one exception: for Fanar on Moroccan Arabic, annotators score steered fluency and coherence below the unsteered baseline, so the human average falls, where the judge average rises. Together, these results support a key interpretability finding: \emph{dialectal behavior is partly localized in identifiable neurons, but robust control requires the broader distributed direction captured by vector steering.} This conclusion
also holds after removing any risk of overlap between the MADAR data
used for steering extraction and the AL-QASIDA evaluation set.
More fine-grained results are provided in Appendices~\ref{app:judge_detailed} and~\ref{app:human-eval}, with qualitative generation examples in Tables~\ref{tab:qualitative_allam} and~\ref{tab:qualitative_fanar}.
\paragraph{Statistical significance:}
Prompt-paired Wilcoxon tests show that 96/120 LLM-judge comparisons remain significant after Holm--Bonferroni correction. Vector steering significantly outperforms neuron steering in 38/40 tests and the unsteered baseline in 31/40, with dialect authenticity and MSA formality significant across all eight model--dialect groups ($|r_{\mathrm{rb}}|=0.43$--$0.99$ for vector--neuron). In contrast, only 7/27 significant neuron--unsteered differences favor neuron steering. Full results, including 10,000-bootstrap confidence intervals and effect sizes, are reported in Appendix~\ref{app:judge-significance}.

\subsection{MSA Prompts}
We evaluate on 300 MSA prompt synthetic samples, steering each toward the target dialects. Neuron steering failed entirely, with dialect authenticity collapsing to the minimum judge score and ADI2 reaching zero, confirming that sparse interventions can reinforce but not induce dialectal generation. Only vector steering results are therefore reported in Table~\ref{tab:msa2dialect_summary}.

Vector steering produces dialectal signal across all targets, though performance is lower than in the mono-dialect setting with ALLaM being more responsive than Fanar. Crucially, the fact that models produce dialectal output despite receiving MSA prompts shows that vector steering exerts influence at the activation level, overriding the prompt's linguistic register. These results reinforce the mono-dialect finding that distributed residual-space directions are more effective than sparse neuron interventions.

\begin{table}[t]
\centering
\scriptsize
\setlength{\tabcolsep}{3pt}
\begin{tabular*}{\columnwidth}{@{\extracolsep{\fill}}lcccccc@{}}
\toprule
& \multicolumn{3}{c}{ALLaM-7B-Instruct-preview} & \multicolumn{3}{c}{Fanar-1-9B-Instruct} \\
\cmidrule(lr){2-4}\cmidrule(lr){5-7}
Dialect & ADI2 & macro & Judge & ADI2 & macro & Judge \\
\midrule
EGY & 0.215 & 0.221 & 4.523 & 0.059 & 0.066 & 3.884 \\
LEB & 0.108 & 0.291 & 4.014 & 0.008 & 0.047 & 3.511 \\
SYR & 0.080 & 0.219 & 4.143 & 0.004 & 0.023 & 3.620 \\
SAU & 0.046 & 0.102 & 3.918 & 0.000 & 0.003 & 3.611 \\
QAT & 0.069 & 0.158 & 3.676 & 0.002 & 0.017 & 3.522 \\
MOR & 0.293 & 0.350 & 4.321 & 0.102 & 0.139 & 3.640 \\
\bottomrule
\end{tabular*}
\caption{
MSA-to-dialect vector-steering results using layer 20 and 30 steered tokens.
Higher is better for all metrics. Judge denotes the average LLM-as-a-judge
score. EGY, LEB, SYR, SAU, QAT, and MOR denote Egyptian, Lebanese, Syrian,
Saudi, Qatari, and Moroccan Arabic.
}
\label{tab:msa2dialect_summary}
\end{table}

\section{ Analysis and Discussion}
\label{sec:analysis}

To explain the gap between neuron and vector steering, we compare the residual
dialect directions used by vector steering with the subspace spanned by the
down-projection directions of the LAPE-selected MLP neurons. This yields a
residual-subspace coverage score measuring how much of each dialect vector is
captured by the sparse neuron set. Formal details and random-baseline tests are
provided in Appendices~\ref{app:neuron-vector-coverage-formalism} and
\ref{app:neuron-vector-coverage-random}.

\begin{figure}[t]
    \centering
    \includegraphics[width=\columnwidth]{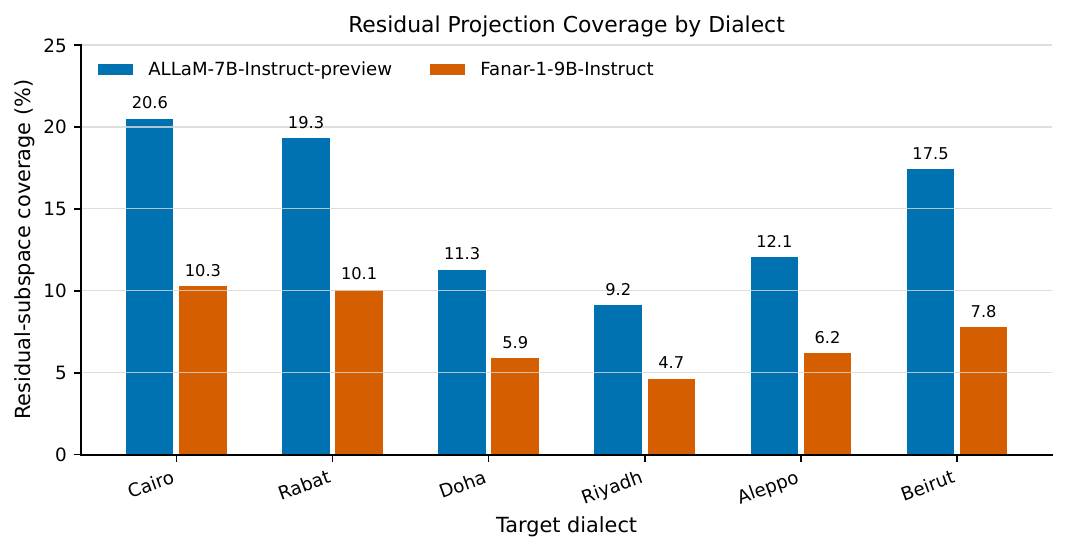}

    \vspace{0.4em}

    \includegraphics[width=\columnwidth]{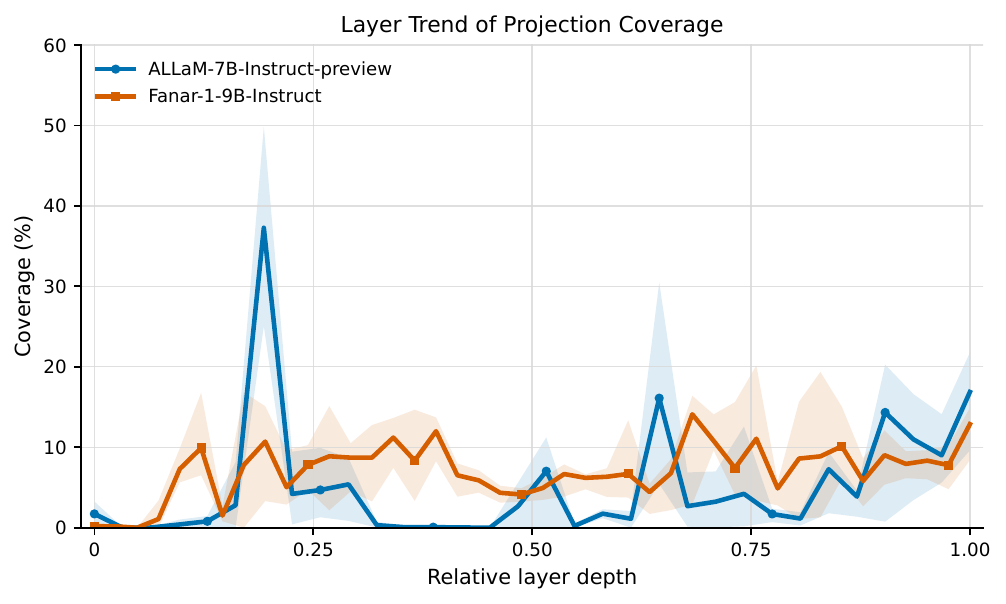}

\caption{
Residual-subspace coverage of vector-steering directions by LAPE-selected
neuron output subspaces. Top: aggregate coverage by dialect. Bottom: mean
coverage across normalized layer depth, with shaded variation across dialects.
Higher values indicate stronger overlap between sparse neuron subspaces and
distributed dialect directions.
}
    \label{fig:residual-projection-coverage}
\end{figure}

Figure~\ref{fig:residual-projection-coverage} shows that LAPE-selected neurons
capture a meaningful but incomplete projection of the residual dialect
directions. Although the selected neurons occupy less than 1\% of MLP
intermediate dimensions, they explain a non-trivial portion of the dialect
direction. However, most of the residual
direction remains outside this sparse subspace. This explains the main empirical
gap: neuron steering provides localized interpretability, while vector steering
better captures the distributed activation shift needed for robust dialect
control.

\section{Related Work}
\label{sec:relatedwork}
Our work on dialectal Arabic control in LLMs sits at the intersection of Arabic NLP and mechanistic interpretability.

\paragraph{Evaluation of Arabic linguistic variation:}
LLMs remain weak on dialectal Arabic, often defaulting to MSA rather than
generating dialectal content \citep{robinson-etal-2025-al}. Prior work analyzes
Arabic model internals across varieties \citep{abdelali-etal-2022-post} and
shows that MSA-to-dialect transfer is uneven, geographically shaped, and prone
to negative interference \citep{khalak-etal-2026-fusha}. Relatedly, \citet{elshabrawy2025alignmenthurtsdecouplingrepresentational} find that excessive entanglement with MSA
can hurt dialect generation and use subspace interventions to decouple dialect
modeling from the dominant standard variety. Benchmarks such as ARADiCE further
enable fine-grained evaluation of dialectal and cultural capabilities
\citep{mousi-etal-2025-aradice}.

\paragraph{Neuron-level language control and activation steering:}
\citet{tang-etal-2024-language} identify language-specific neurons with LAPE and
show that activating them can steer output language. Subsequent work shows that amplifying such neurons improves steering \citep{rahmanisa2025language_specific}. Language arithmetic extends LAPE-based analysis by systematically
activating or suppressing language neurons and showing overlap among related
languages \citep{gurgurov-etal-2025-language}.

Prior work has shown that linguistic properties can be localized in individual neurons and that fine-tuning reorganizes their internal representations \citep{durrani-etal-2020-analyzing,durrani-etal-2022-latent}. A complementary line of work has established that high-level concepts can be extracted as linear directions in activation space and used to steer model behavior at inference time. \citet{turner2024activation} introduced activation addition, showing that adding a contrastive steering vector to residual-stream activations reliably shifts model behavior toward a target concept. \citet{panickssery2024steering} found that mean differences between contrastive activations at the last token position consistently outperform other extraction strategies. \citet{zou2025representation} demonstrated that traits such as honesty and toxicity can be identified and manipulated via linear directions. \citet{chen2025persona} extended this paradigm to personality traits, showing that persona vectors extracted via contrastive prompting can monitor and control sycophancy, hallucination, and malicious behavior, and that activations from response tokens yield more effective steering directions than prompt tokens. \citet{aneja2026intrinsicguardrailssemanticgeometry} show that personality-related
semantic directions can transfer across model variants and regulate emergent
misalignment.

Unlike prior work on cross-lingual transfer, we operate within a single language family where dialectal boundaries are fuzzy and MSA overlap is high. We show that Arabic LLMs nonetheless encode dialect identity as structured geometric properties of their hidden states, and that contrastive activation directions can steer generation toward a target dialect at inference time, revealing intra-linguistic variation as an underexplored dimension of LLM representation space.




\section{Conclusion and Future Work}
\label{sec:conclusion}

We explored the mechanistic interpretability of Arabic dialect encoding in
LLMs, finding that dialectal information is neither fully localized nor
entirely diffuse, with dialect-associated neurons concentrating in late
layers yet capturing only a partial projection of the broader
residual-space dialect direction. These neurons occupy under 1\% of MLP
intermediate dimensions and separate MSA from the spoken dialects far more
sharply than the dialects separate from one another. We applied both
neuron-based and distributed vector steering, with vector steering providing
more reliable dialect control across both models. 

Vector steering is also
the only method that induces dialect from MSA prompts: neuron steering fails
outright there, with dialect authenticity collapsing to the minimum score,
confirming that sparse interventions can reinforce a dialect but not induce
one. Both interventions leave model weights unchanged and add no meaningful
inference cost. The gains are not uniform: for Fanar on Moroccan Arabic,
annotators judge steered output less fluent and less coherent than the
baseline despite a clear authenticity gain, a trade-off the LLM judge does
not register.

Future work should extend this analysis to additional dialects and
mixed-dialect settings, explore adaptive and compositional steering methods,
and investigate how dialect representations interact with other stylistic
and sociolinguistic attributes in multilingual and multimodal models. Our
weakest results come from the dialects with the smallest extraction corpora,
suggesting that low-resource varieties may require extraction strategies
that do not depend on large parallel sets.The high neuron overlap among
regionally proximate dialects points to reusable subcircuits, raising the
question of whether steering directions can transfer within a region rather
than being estimated per dialect.
Finally, dialect evaluation needs metrics that score output quality
alongside dialectness, since both ADI2 and LLM judges have blind spots that
only human annotation exposed.

\section*{Limitations}

Our experiments focus on a limited set of Arabic dialects and two Arabic-centric LLMs, and results may not fully generalize to other dialects, architectures, or multilingual models. In addition, evaluation of dialectal generation remains inherently challenging: automatic metrics and LLM-as-a-judge scores capture complementary aspects of performance, but neither provides a complete measure of dialect authenticity or sociolinguistic naturalness. Human evaluations are inherently subjective, and inter-annotator agreement, while reasonable, reflects the ambiguity inherent in dialect perception. Finally, while our analysis identifies interpretable dialect-related structures, the relationship between localized neuron representations and distributed residual-space features remains only partially understood, and further work is needed to characterize these mechanisms more precisely.

\section*{Ethics and Broader Impact}

This work aims to improve dialectal inclusivity in Arabic LLMs by reducing the strong bias toward Modern Standard Arabic and enabling more natural generation for underrepresented dialect communities. We focus on inference-time steering methods that do not require retraining or modification of model parameters, which may lower the computational cost of dialect adaptation and improve accessibility for low-resource varieties.

At the same time, dialect steering technologies may be misused to imitate regional linguistic styles in deceptive or manipulative contexts. Our methods are intended for research on controllable and interpretable language generation, not for impersonation or misinformation. In addition, Arabic dialect boundaries are inherently fluid and socially complex, and automatic evaluations may not fully capture sociolinguistic authenticity or community perceptions of dialect use. We therefore encourage future work incorporating broader human evaluation across speaker communities and dialect backgrounds.

Annotations were annotated by trained Arabic annotators recruited through a thirdparty company and compensated at the standard hourly rate for their location. Annotators signed non-disclosure agreements prior to participation. We provide clear task instructions, avoid collecting annotator personal data beyond what is required for payment and administration by the vendor, and restrict dataset release to permitted uses consistent with the underlying licenses.


\bibliography{custom}

\clearpage
\appendix
\section*{Appendix}

\section{Neuron-Based Steering Ablations}
\label{app:neuron-steer-abls}

We conducted ablation experiments to assess the contribution of each neuron-steering parameter and to select an effective steering configuration. Specifically, we varied the target-dialect amplification factor $\alpha$, the MSA suppression factor $\gamma$, and the competitor-dialect suppression factor $\gamma_{\mathrm{comp}}$. In each ablation, one parameter was varied while the remaining two were held fixed, allowing us to isolate the effect of each component.

Experiments were conducted using ALLaM-7B-Instruct-preview and Fanar-1-9B-Instruct, with Egyptian Arabic and Moroccan Arabic as target dialects.

\subsection{Target Dialect Coefficient Ablations}
\begin{figure*}[t]
    \centering

    \begin{subfigure}[t]{0.49\textwidth}
        \centering
        \includegraphics[width=\linewidth]{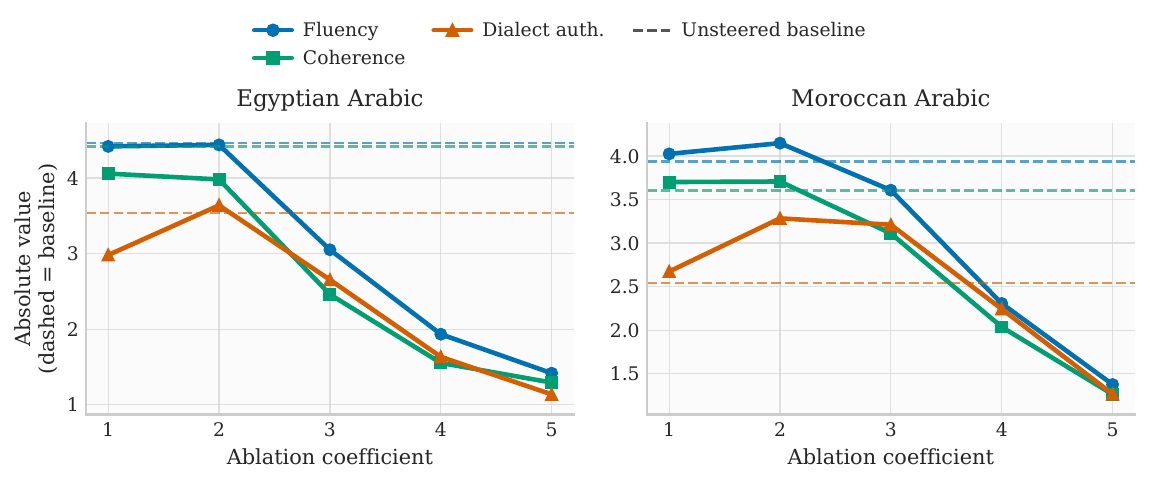}
        \caption{ALLaM-7B-Instruct-preview LLM-as-a-judge scores.}
        \label{fig:allam-neuron-judge}
    \end{subfigure}
    \hfill
    \begin{subfigure}[t]{0.49\textwidth}
        \centering
        \includegraphics[width=\linewidth]{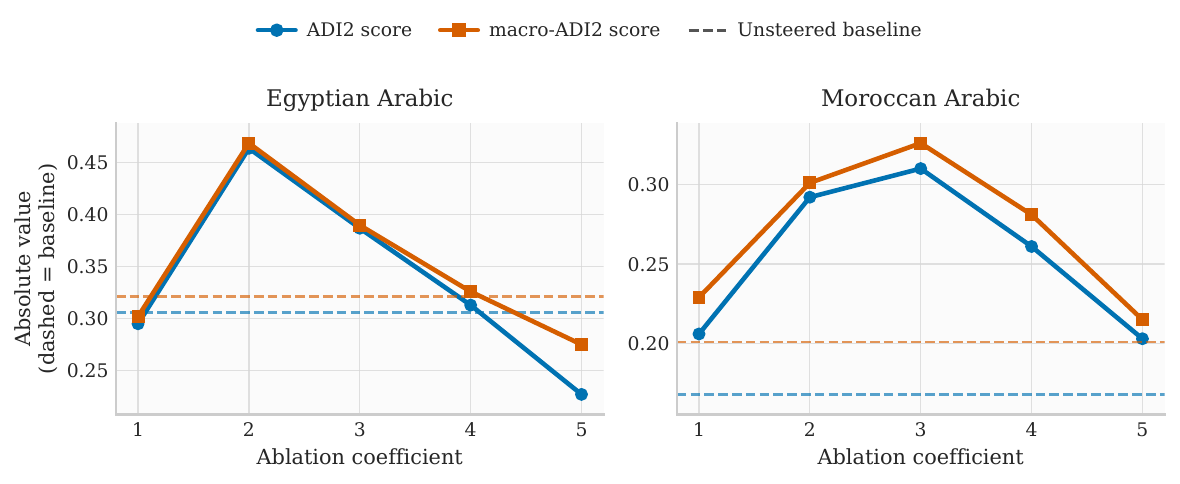}
        \caption{ALLaM-7B-Instruct-preview ADI2 scores.}
        \label{fig:allam-neuron-macro}
    \end{subfigure}

    \vspace{0.7em}

    \begin{subfigure}[t]{0.49\textwidth}
        \centering
        \includegraphics[width=\linewidth]{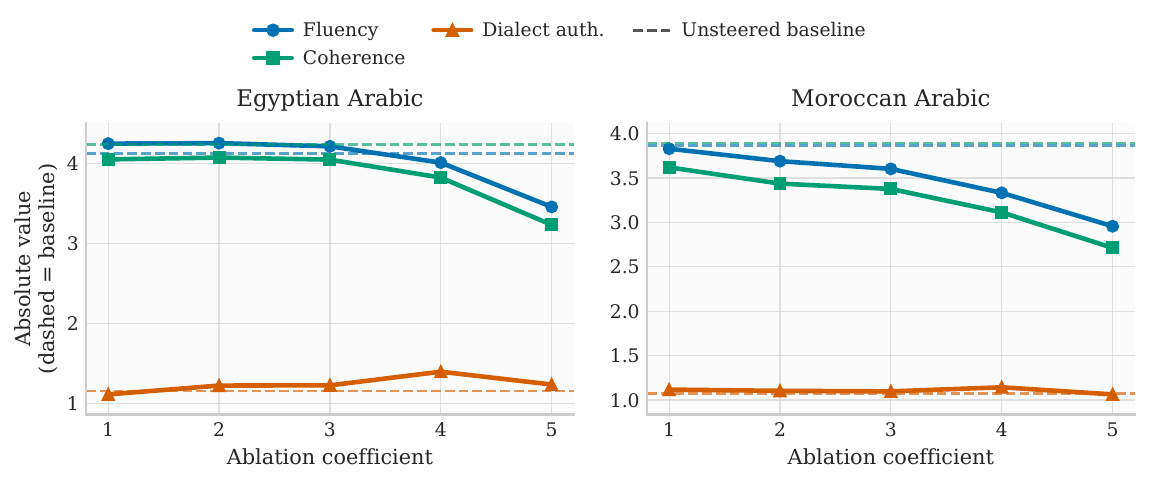}
        \caption{Fanar-1-9B-Instruct LLM-as-a-judge scores.}
        \label{fig:fanar-neuron-judge}
    \end{subfigure}
    \hfill
    \begin{subfigure}[t]{0.49\textwidth}
        \centering
        \includegraphics[width=\linewidth]{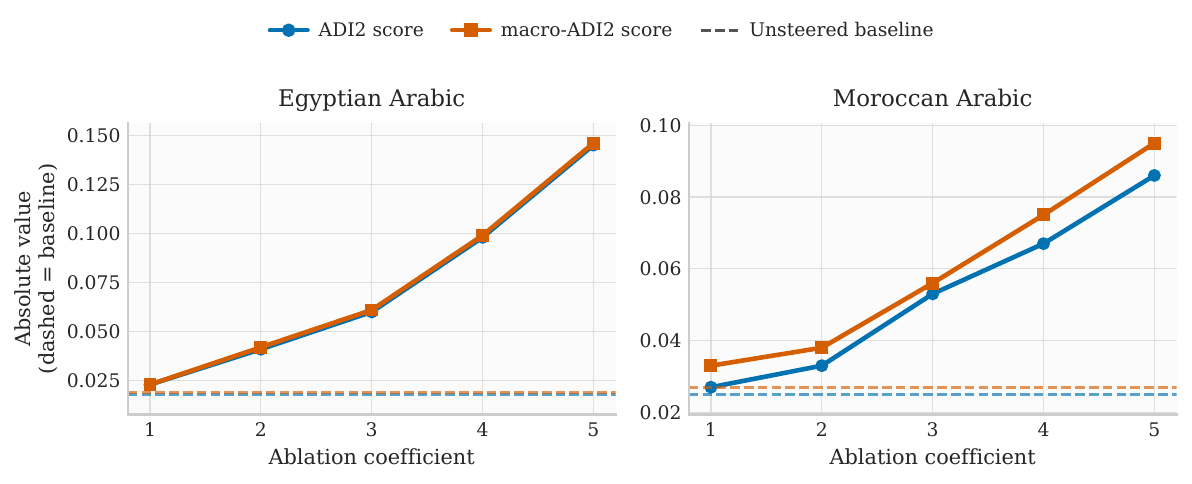}
        \caption{Fanar-1-9B-Instruct ADI2 scores.}
        \label{fig:fanar-neuron-macro}
    \end{subfigure}

    \caption{Neuron-steering target-dialect amplification factor ablation results for \textbf{ALLaM-7B-Instruct-preview} and \textbf{Fanar-1-9B-Instruct}. For each model, the left plot reports LLM-as-a-judge scores for fluency, coherence, and dialect authenticity, while the right plot reports automatic dialect metrics using ADI2  scores. Results are shown across steering coefficients for Egyptian Arabic and Moroccan Arabic, with dashed lines indicating the unsteered baseline.}
    \label{fig:neuron-steering-coeff-ablation-results}
\end{figure*}

For this ablation, we vary the target-dialect amplification coefficient
$\alpha$ while fixing the MSA suppression coefficient and the non-target
competitor suppression coefficient to $\gamma=\gamma_{\mathrm{comp}}=0.9$.
This isolates the effect of strengthening target-dialect neurons during
decoding. As shown in Figure~\ref{fig:neuron-steering-coeff-ablation-results},
ALLaM exhibits a non-monotonic response to $\alpha$. Moderate amplification
improves dialectal performance, with $\alpha=2.0$ providing the best overall
trade-off across Egyptian Arabic and Moroccan Arabic. Larger values lead to a
sharp degradation in LLM-as-a-judge scores, particularly fluency and coherence,
suggesting that excessive amplification can distort generation quality even
when dialectal signals are strengthened.

For Fanar, increasing $\alpha$ produces stronger gains in the automatic
dialectal metrics, especially ADI2  scores, which improve consistently
as the coefficient increases. However, the LLM-as-a-judge results show that very
large amplification reduces fluency and coherence, while dialect authenticity
improves only modestly. We therefore select $\alpha=4.0$ for Fanar as a
balanced setting: it substantially improves dialectal metrics while avoiding
the larger quality degradation observed at $\alpha=5.0$. Overall, these
results indicate that the target-dialect amplification coefficient is a main
driver of neuron-steering performance, but its optimal value is model-dependent.

\subsection{MSA Neuron Suppression Ablations}

\begin{figure*}[h!]
    \centering

    \begin{subfigure}[t]{0.49\textwidth}
        \centering
        \includegraphics[width=\linewidth]{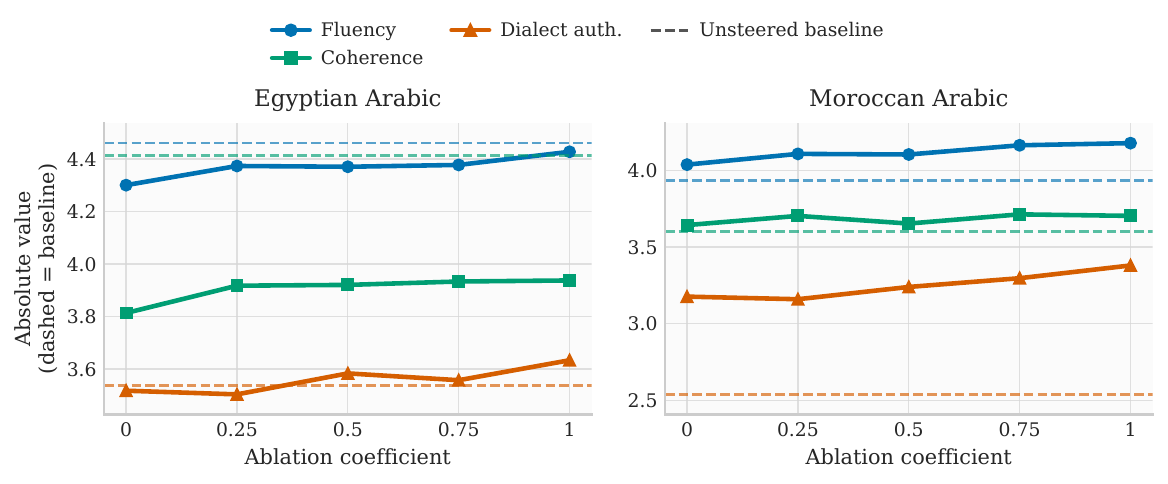}
        \caption{ALLaM-7B-Instruct-preview LLM-as-a-judge scores.}
        \label{fig:allam-neuron-msa-supp-judge}
    \end{subfigure}
    \hfill
    \begin{subfigure}[t]{0.49\textwidth}
        \centering
        \includegraphics[width=\linewidth]{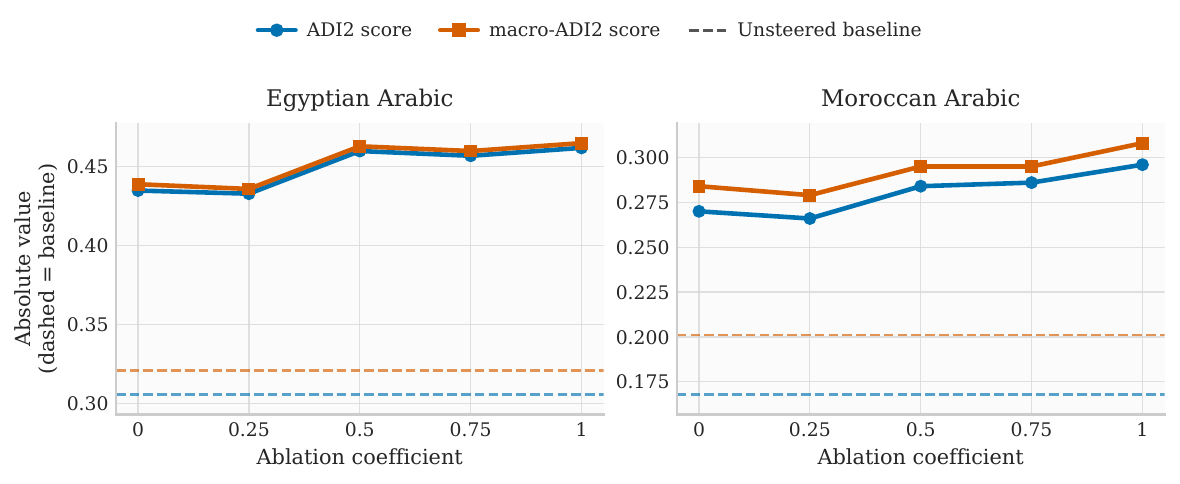}
        \caption{ALLaM-7B-Instruct-preview ADI2 scores.}
        \label{fig:allam-neuron-msa-supp-macro}
    \end{subfigure}

    \vspace{0.7em}

    \begin{subfigure}[t]{0.49\textwidth}
        \centering
        \includegraphics[width=\linewidth]{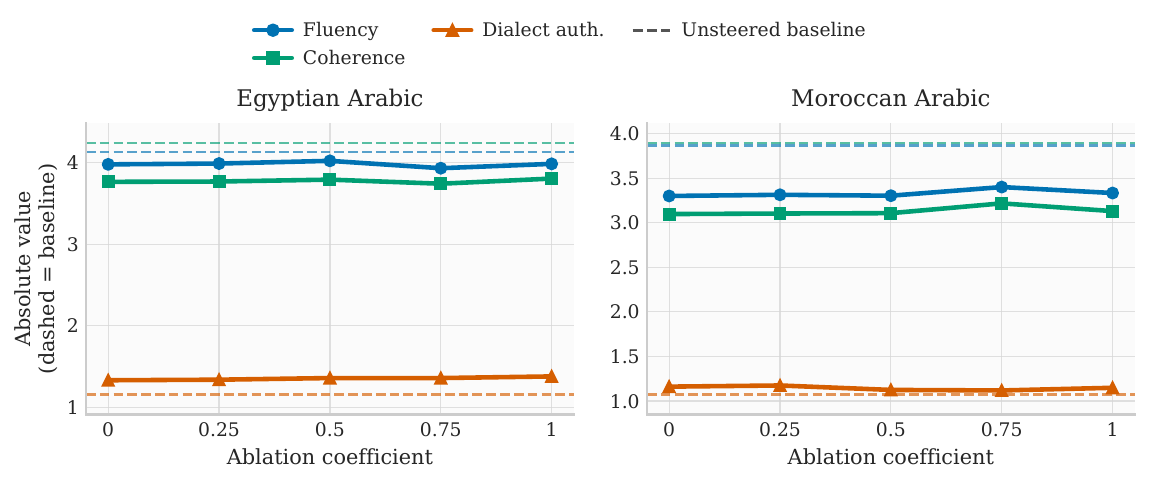}
        \caption{Fanar-1-9B-Instruct LLM-as-a-judge scores.}
        \label{fig:fanar-neuron-msa-supp-judge}
    \end{subfigure}
    \hfill
    \begin{subfigure}[t]{0.49\textwidth}
        \centering
        \includegraphics[width=\linewidth]{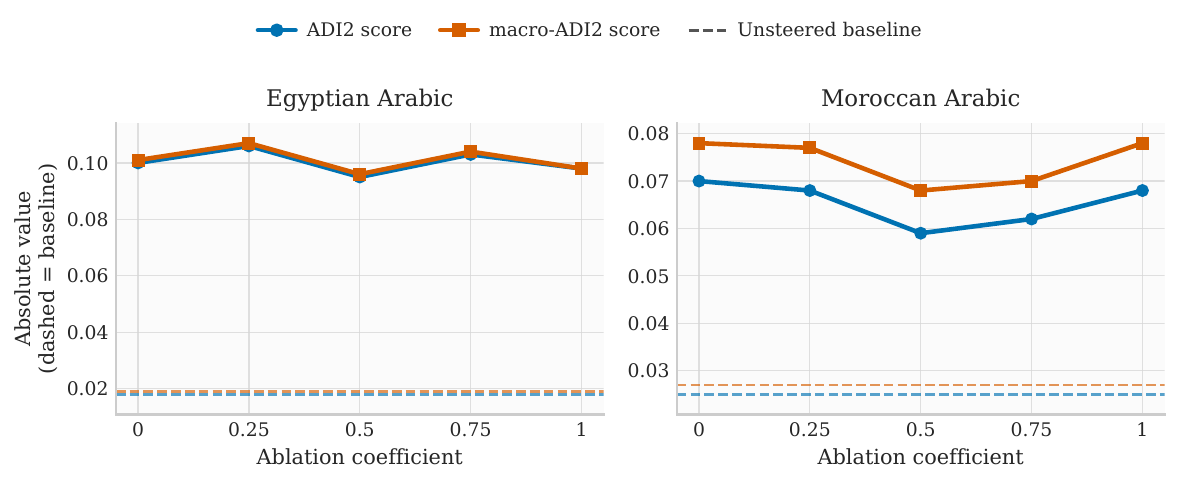}
        \caption{Fanar-1-9B-Instruct ADI2 scores.}
        \label{fig:fanar-neuron-msa-supp-macro}
    \end{subfigure}

    \caption{Neuron-steering MSA suppression factor ablation results for \textbf{ALLaM-7B-Instruct-preview} and \textbf{Fanar-1-9B-Instruct}. For each model, the left plot reports LLM-as-a-judge scores for fluency, coherence, and dialect authenticity, while the right plot reports automatic dialect metrics using ADI2  scores. Results are shown across MSA suppression coefficients for Egyptian Arabic and Moroccan Arabic, with dashed lines indicating the unsteered baseline.}
    \label{fig:neuron-steering-msa-supp-ablation-results}
\end{figure*}

In this ablation, we vary the MSA suppression coefficient $\gamma$ while fixing
the target-dialect amplification coefficient to the best value selected in the
previous ablation, namely $\alpha=2.0$ for ALLaM and $\alpha=4.0$ for Fanar.
We also fix the competitor-dialect suppression coefficient to
$\gamma_{\mathrm{comp}}=0.9$. This setup isolates the effect of suppressing
MSA-specific neurons. As shown in
Figure~\ref{fig:neuron-steering-msa-supp-ablation-results}, varying $\gamma$
has a weaker and less consistent effect than varying the target-dialect
amplification coefficient. For ALLaM, weaker MSA suppression, represented by
larger values of $\gamma$, generally preserves or improves performance, while
strong suppression does not provide clear gains and can reduce fluency and
coherence, especially for Egyptian Arabic.

For Fanar, the effect of $\gamma$ is also limited and non-monotonic. Although
automatic dialect scores remain above the unsteered baseline across most
settings, no single MSA suppression value yields consistent improvements across
both dialects and evaluation metrics. The LLM-as-a-judge scores further indicate
that suppressing MSA neurons does not reliably improve dialect authenticity and
may coincide with lower fluency and coherence. Overall, the results suggest that
MSA-neuron suppression is not a driver of neuron-steering performance.
The gains are primarily attributable to target-dialect neuron amplification,
while aggressive MSA suppression can be unnecessary or harmful.

\subsection{Non-Target Dialects Neuron Ablations}
In this ablation, we vary the non-target dialect suppression coefficient
$\gamma_{\mathrm{comp}}$ while fixing the target-dialect amplification
coefficient to the best value selected earlier, namely $\alpha=2.0$ for ALLaM
and $\alpha=4.0$ for Fanar. We also fix the MSA suppression coefficient to
$\gamma=1.0$. Since $\gamma_{\mathrm{comp}}$ is a multiplicative suppression
factor, smaller values correspond to stronger suppression, while
$\gamma_{\mathrm{comp}}=1$ corresponds to no suppression of non-target dialect
neurons.

\begin{figure*}[h!]
    \centering

    \begin{subfigure}[t]{0.49\textwidth}
        \centering
        \includegraphics[width=\linewidth]{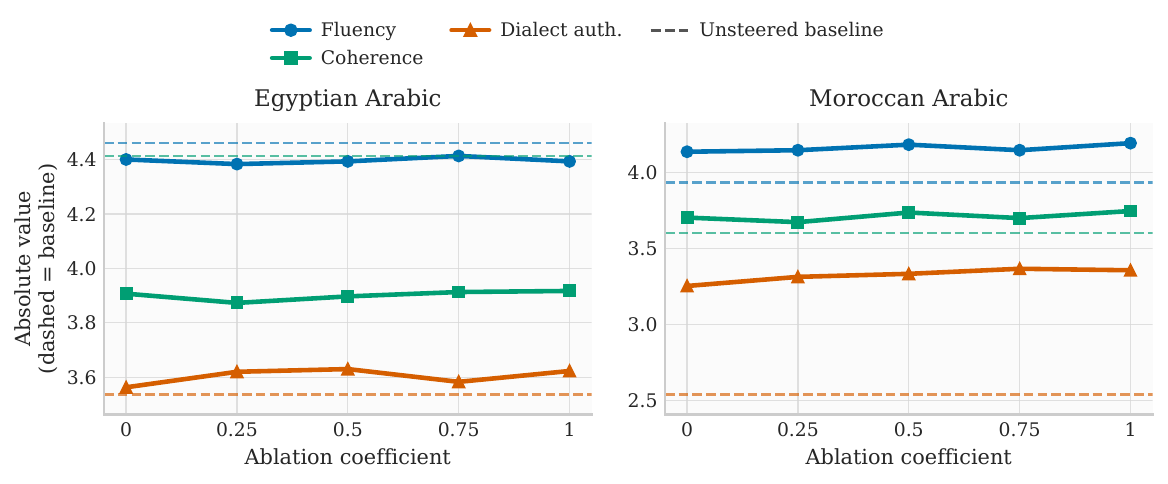}
        \caption{ALLaM-7B-Instruct-preview LLM-as-a-judge scores.}
        \label{fig:allam-neuron-non-target-supp-judge}
    \end{subfigure}
    \hfill
    \begin{subfigure}[t]{0.49\textwidth}
        \centering
        \includegraphics[width=\linewidth]{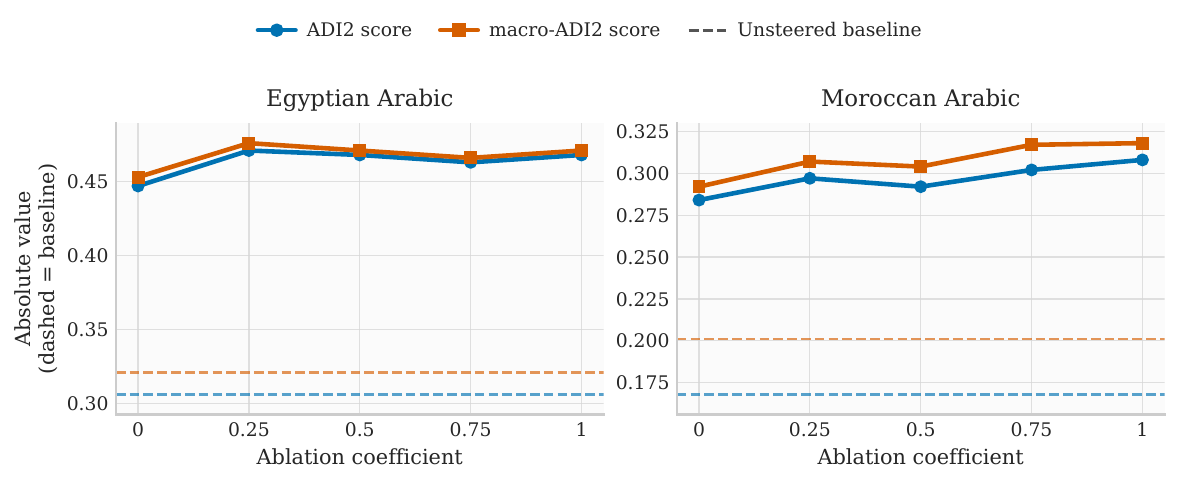}
        \caption{ALLaM-7B-Instruct-preview ADI2 scores.}
        \label{fig:allam-neuron-non-target-supp-macro}
    \end{subfigure}

    \vspace{0.7em}

    \begin{subfigure}[t]{0.49\textwidth}
        \centering
        \includegraphics[width=\linewidth]{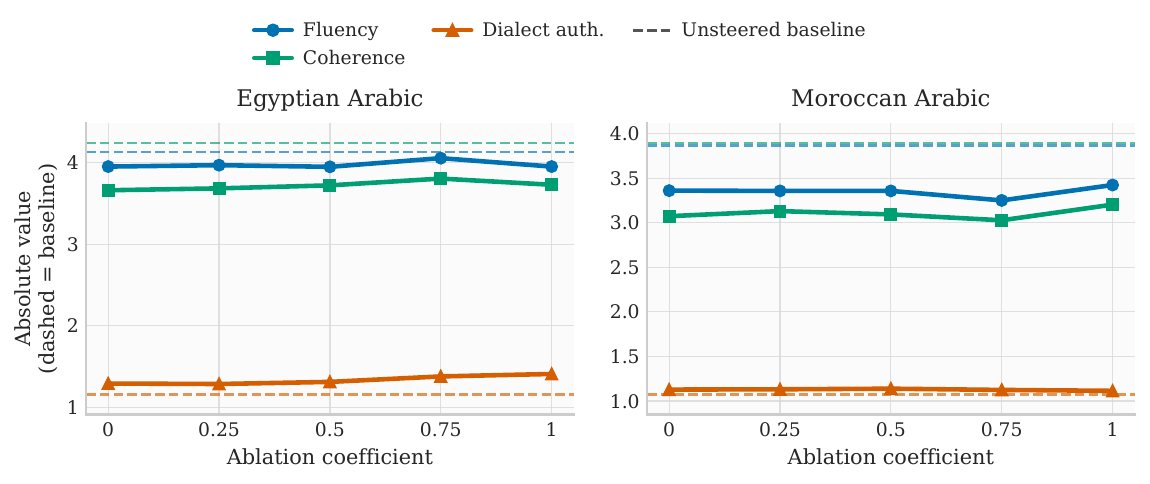}
        \caption{Fanar-1-9B-Instruct LLM-as-a-judge scores.}
        \label{fig:fanar-neuron-non-target-supp-judge}
    \end{subfigure}
    \hfill
    \begin{subfigure}[t]{0.49\textwidth}
        \centering
        \includegraphics[width=\linewidth]{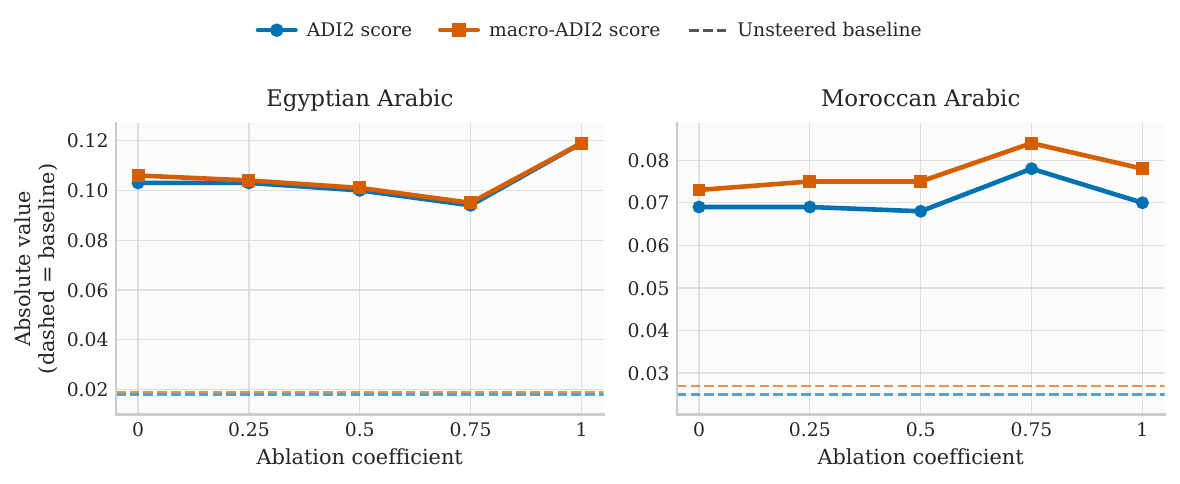}
        \caption{Fanar-1-9B-Instruct ADI2 scores.}
        \label{fig:fanar-neuron-non-target-supp-macro}
    \end{subfigure}

    \caption{Neuron-steering non-target dialect suppression factor ablation results for \textbf{ALLaM-7B-Instruct-preview} and \textbf{Fanar-1-9B-Instruct}. For each model, the left plot reports LLM-as-a-judge scores for fluency, coherence, and dialect authenticity, while the right plot reports automatic dialect metrics using ADI2  scores. Results are shown across non-target dialect suppression coefficients for Egyptian Arabic and Moroccan Arabic, with dashed lines indicating the unsteered baseline.}
    \label{fig:neuron-steering-non-target-supp-ablation-results}
\end{figure*}

As shown in
Figure~\ref{fig:neuron-steering-non-target-supp-ablation-results}, varying
$\gamma_{\mathrm{comp}}$ has a limited and inconsistent effect compared with
target-dialect amplification. For ALLaM, automatic dialect scores remain above
the unsteered baseline across settings, but stronger non-target suppression
does not consistently improve LLM-as-a-judge scores. For Fanar, automatic scores
also remain above baseline, yet the LLM-as-a-judge results show no reliable gains in
fluency, coherence, or dialect authenticity. These results suggest that
suppressing non-target dialect neurons is not a primary driver of steering
performance, and that aggressive non-target suppression is generally
unnecessary.

\section{Vector Steering}
\label{app:vector-steer}

\subsection{Analysis of Extracted Dialect Representations}
\label{app:vector_analysis}
\begin{figure}[h!]
\centering
\begin{subfigure}[b]{0.32\textwidth}
\includegraphics[width=\textwidth]{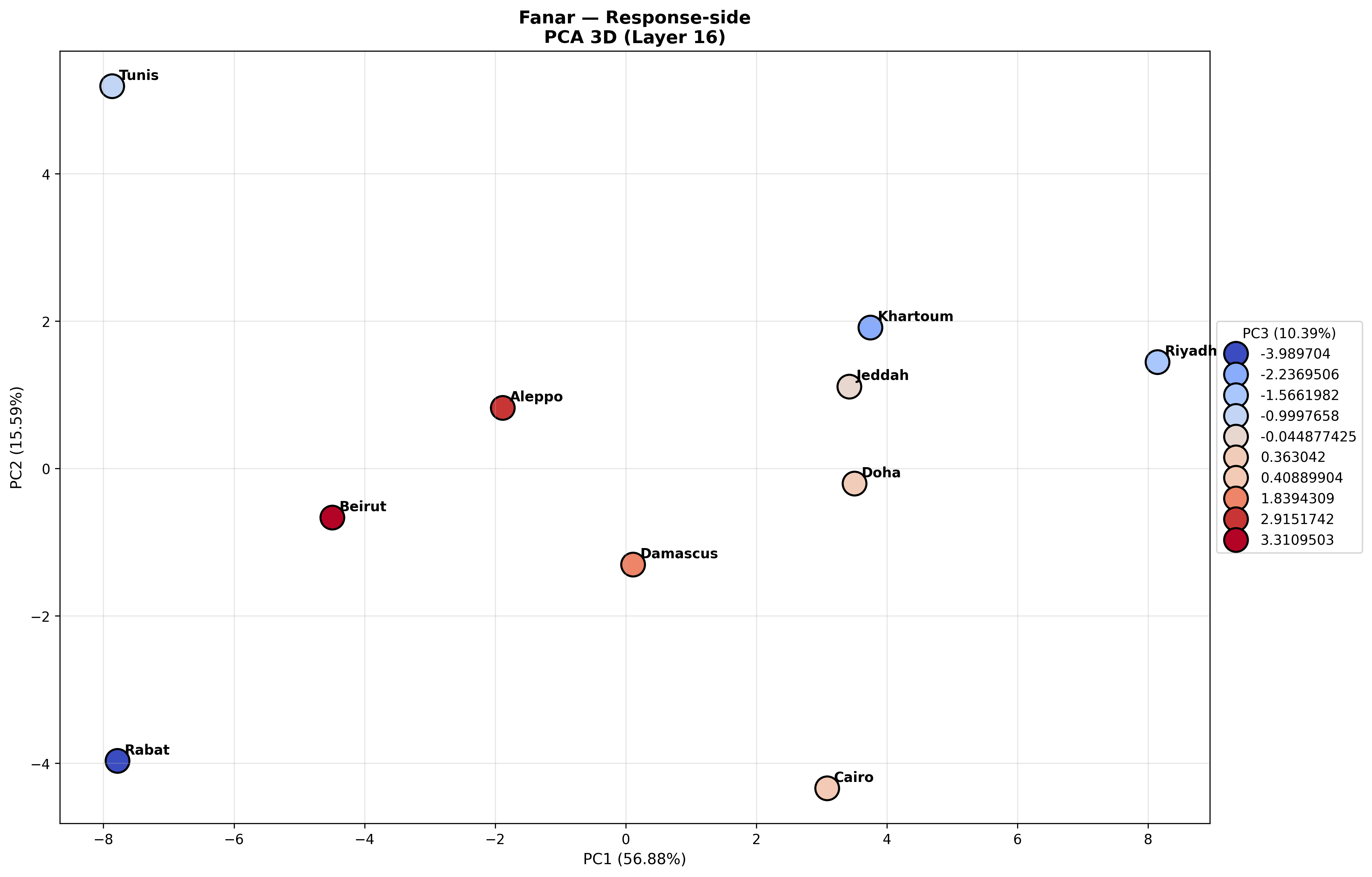}
\caption{Fanar}
\end{subfigure}
\hfill
\begin{subfigure}[b]{0.32\textwidth}
\includegraphics[width=\textwidth]{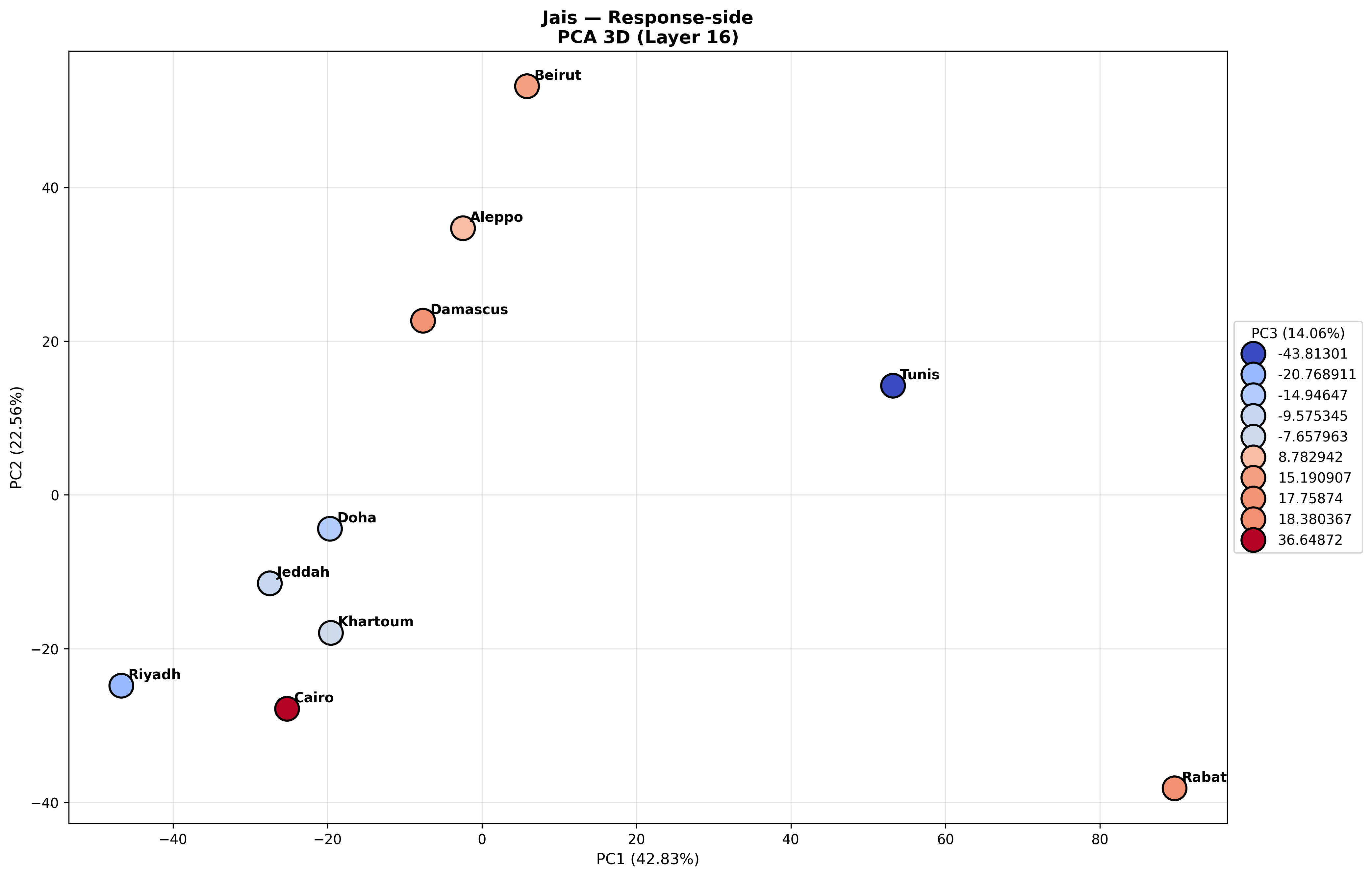}
\caption{Jais}
\end{subfigure}
\hfill
\begin{subfigure}[b]{0.32\textwidth}
\includegraphics[width=\textwidth]{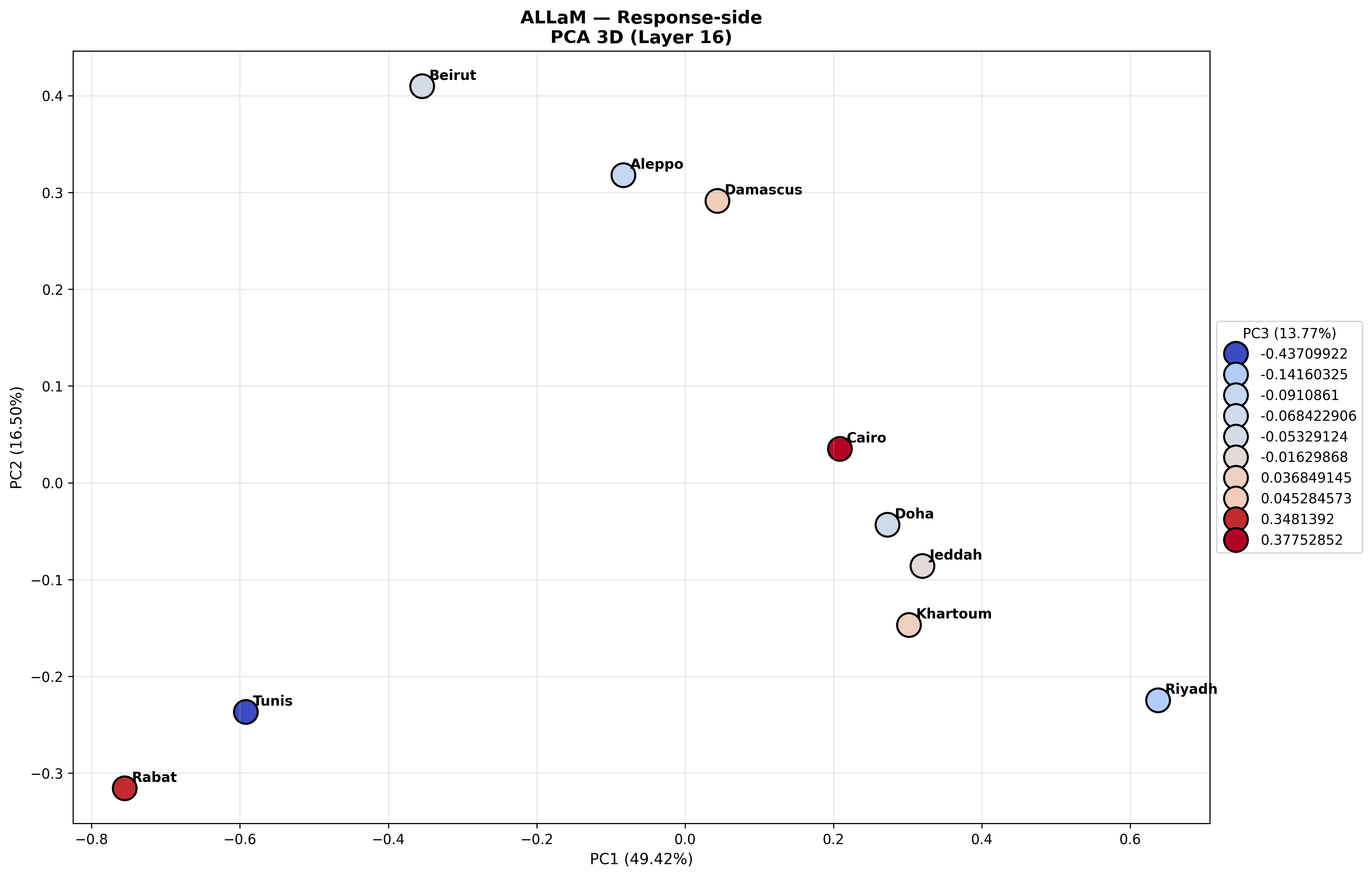}
\caption{ALLaM}
\end{subfigure}
\caption{PC1–PC2 positions with PC3 encoded by color of mean response-side hidden states at layer 16 for Fanar, Jais, and ALLaM. Color encodes the third principal component. Dialects are labeled by city.}
\label{fig:pca_layer16}
\end{figure}

\paragraph{Dialectal Geometry.}
Figure~\ref{fig:pca_layer16} shows PCA projections of the mean response-side hidden states at layer 16 for three Arabic LLMs: ALLaM-7B-Instruct-preview, Fanar-1-9B-Instruct, and Jais-2 \citep{anwar2026jais2}, across ten Arabic dialect cities spanning Maghrebi, Levantine, and Gulf varieties. The first three principal components accounting for the majority of variance (PC1 alone captures 56.88\%,  49.42\% ,and 42.83\% of variance for Fanar, Jais-2 and ALLaM respectively). Across all three models, dialects do not collapse to a single point in this space; instead, they spread across the principal components in patterns that reflect known geographic and typological groupings. 

Geographically close dialects like Rabat and Tunis consistently occupy positions that are well-separated from Gulf dialects (Riyadh, Doha, Jeddah), with Levantine dialects (Beirut, Aleppo, Damascus) occupying an intermediate region. Crucially, cities that are geographically proximate tend to cluster together in the representation space: Beirut and Aleppo appear close across all three models, as do Riyadh and Jeddah, and Rabat and Tunis. This neighborhood-preserving structure suggests that the models have not merely memorized surface-level dialectal markers but have internalized a latent geography of Arabic variation, where representational distance roughly tracks real-world linguistic proximity.

This broad geographic clustering arises without explicit dialectal supervision, emerging instead from the distributional patterns in the training data. Jais exhibits the most pronounced spread along PC1, with Rabat positioned at the extreme positive end and Riyadh at the extreme negative end, indicating that the first principal component roughly tracks a Maghrebi-to-Gulf axis. Fanar produces a comparable geographic structure, though with tighter clustering among Gulf dialects. ALLaM, by contrast, shows a more compressed representation overall, with most dialects occupying a narrower region of PC1, suggesting that its internal geometry treats Arabic varieties as more similar at this layer. 

The third principal component (encoded as color) reveals an additional axis of variation that does not align cleanly with geography, potentially reflecting register or lexical formality rather than regional origin.

\subsection{Coefficient Ablation}
\label{app:coeff_ablation}

\begin{figure}[h!]
    \centering
    \begin{subfigure}{\columnwidth}
        \centering
        \includegraphics[width=\columnwidth]{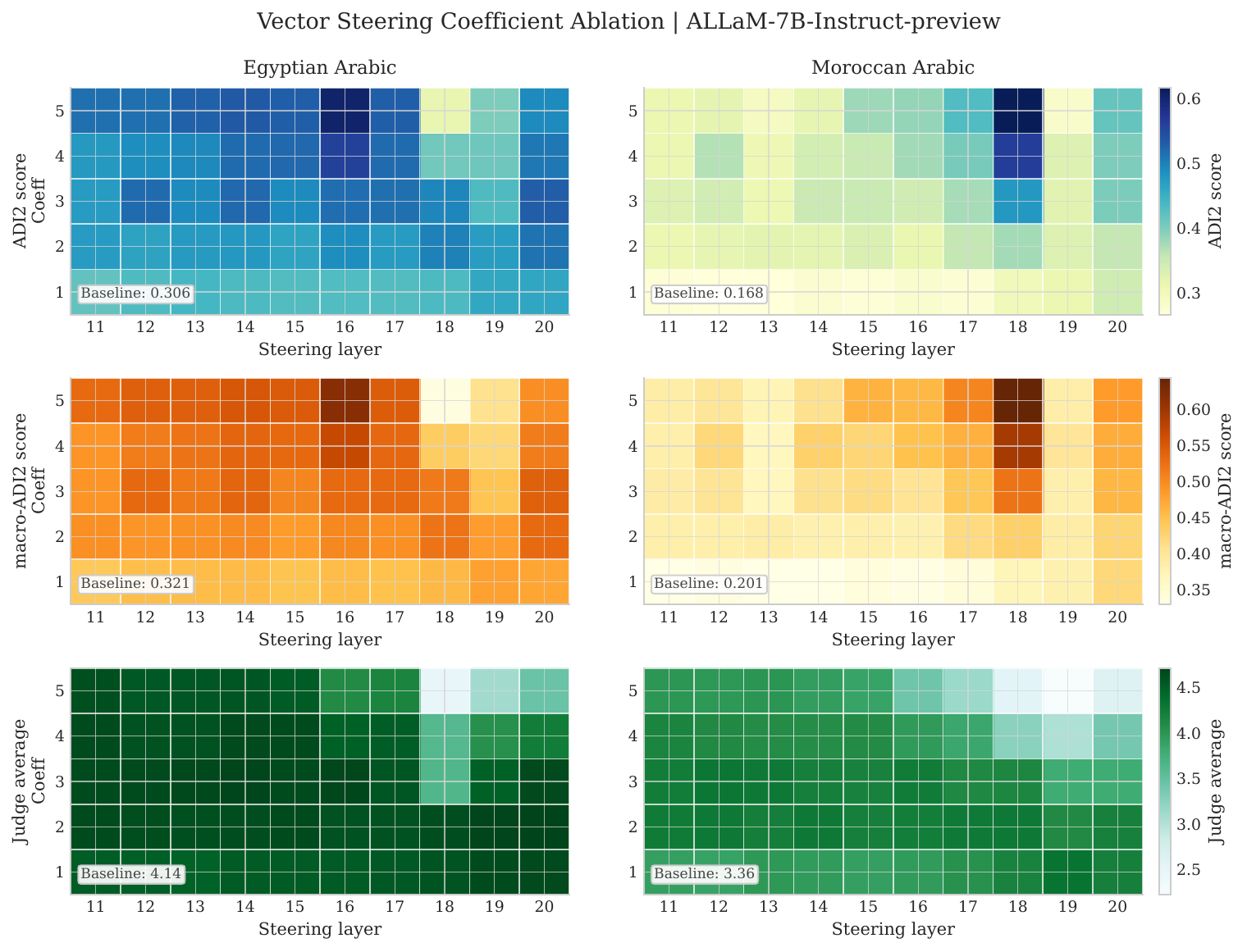}
        \caption{ALLaM-7B-Instruct-preview}
    \end{subfigure}
    \vspace{0.4em}
    \begin{subfigure}{\columnwidth}
        \centering
        \includegraphics[width=\columnwidth]{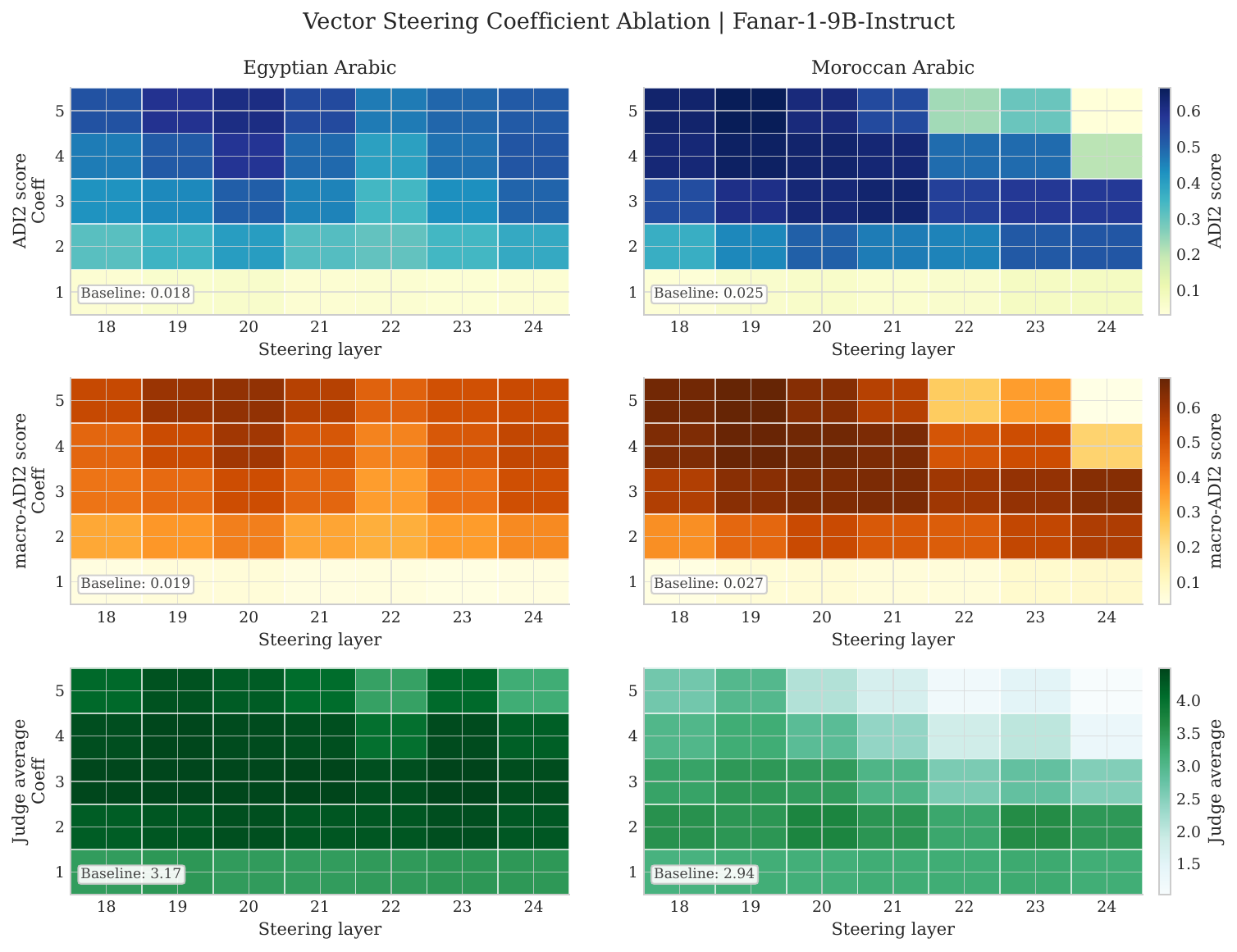}
        \caption{Fanar-1-9B-Instruct}
    \end{subfigure}
    \caption{Steering coefficient ablation heatmaps for ALLaM and Fanar
    across Egyptian and Moroccan Arabic. ADI2, macro, and judge average
    scores are shown per layer--coefficient combination.}
    \label{fig:coeff_ablation}
\end{figure}

Figure~\ref{fig:coeff_ablation} shows the joint layer--coefficient heatmaps for ALLaM and Fanar across Egyptian and Moroccan Arabic. For ALLaM, judge scores remain high across most layer--coefficient combinations for Egyptian Arabic, but begin degrading at later layers (19--21) under higher coefficients, revealing a coherence ceiling that ADI2 alone would not expose: the automatic identifier sees more dialect signal while the judge perceives degrading output quality. Moroccan Arabic shows the same dissociation more sharply, with ADI2 peaking at layer 19 and $\alpha = 5$ while judge quality declines at the same setting. For Fanar, Egyptian Arabic is notably robust: ADI2 and judge scores improve together across the coefficient range, with no clear divergence even at $\alpha = 5$, suggesting that the steering vector aligns well with the model's generative behavior for this dialect. Moroccan Arabic is more sensitive, with judge scores dropping at higher coefficients beyond layer 22 while ADI2 continues to rise, again exposing the benchmark--perception gap.

\subsection{Token Budget Ablation}
\label{app:token_budget}

\begin{figure}[h!]
    \centering
    \begin{subfigure}{\columnwidth}
        \centering
        \includegraphics[width=\columnwidth]{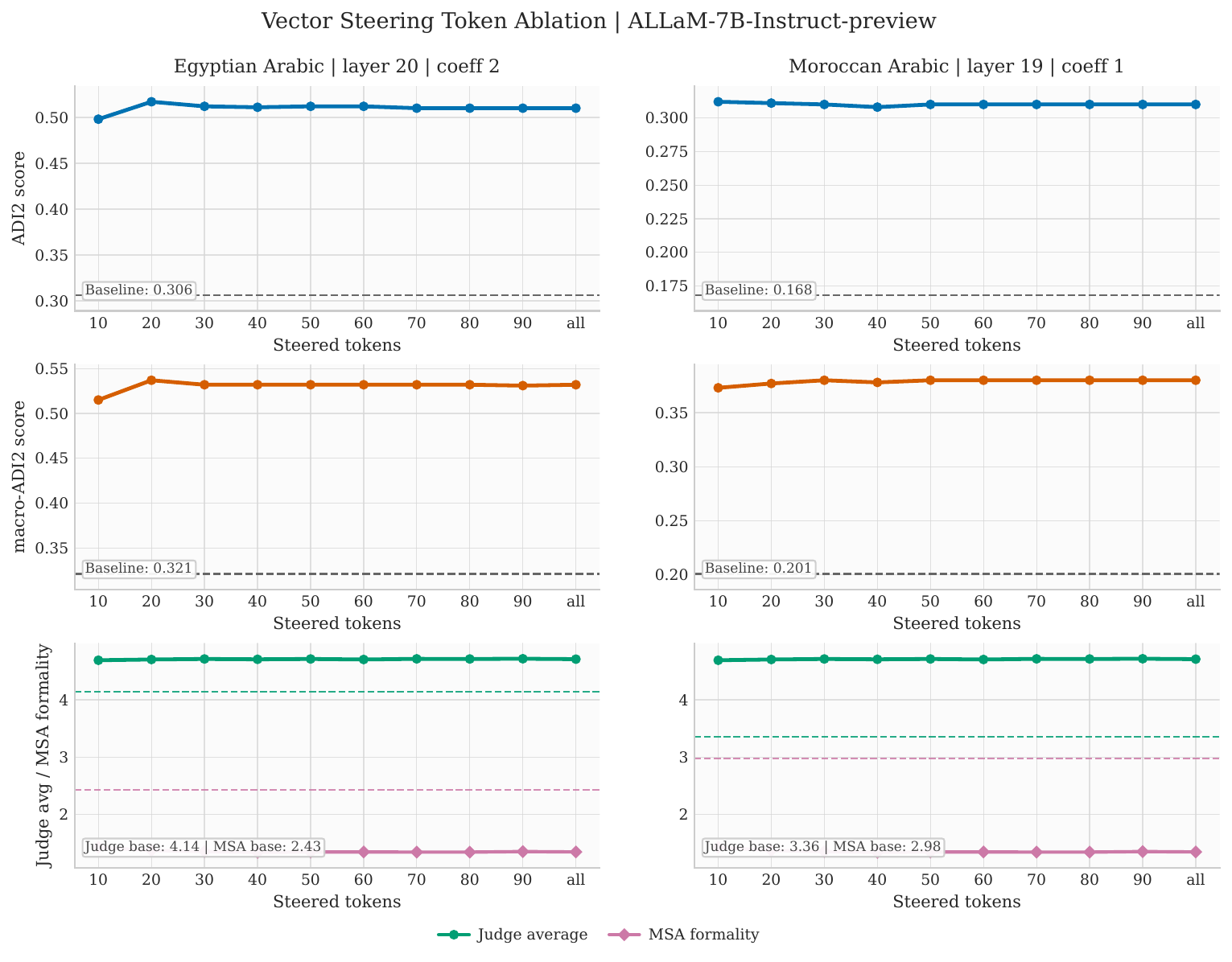}
        \caption{ALLaM-7B-Instruct-preview}
    \end{subfigure}
    \vspace{0.4em}
    \begin{subfigure}{\columnwidth}
        \centering
        \includegraphics[width=\columnwidth]{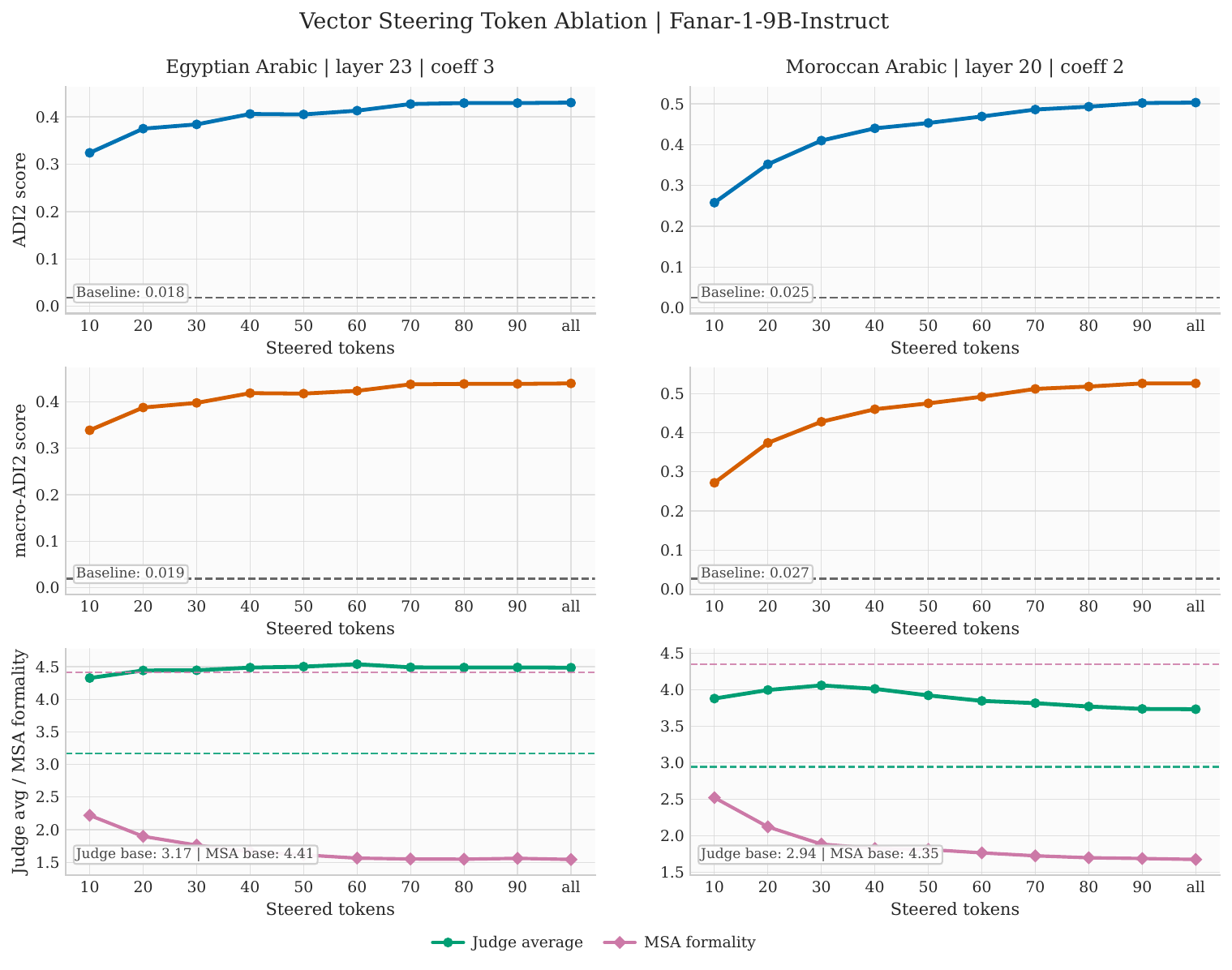}
        \caption{Fanar-1-9B-Instruct}
    \end{subfigure}
    \caption{Token budget ablation for vector steering across Egyptian and
    Moroccan Arabic. Each row shows ADI2 score (top), macro score (middle),
    and judge average with MSA formality (bottom) as a function of the
    number of steered tokens $N$. Dashed lines indicate unsteered baselines.}
    \label{fig:token_budget_ablation}
\end{figure}

Figure~\ref{fig:token_budget_ablation} shows the token budget ablation. For ALLaM, judge scores are flat from $N = 10$ onward for both Egyptian and Moroccan Arabic, and MSA formality stays well below baseline throughout, indicating the model commits to the target dialect after minimal intervention. ADI2 corroborates this, plateauing around 0.51 for Egyptian and 0.31 for Moroccan Arabic after the first 20 tokens. Fanar requires sustained injection to maintain the target register: ADI2 and macro scores rise steadily with $N$ for both dialects. The judge remains stable for Egyptian Arabic across all values of $N$, but shows a gradual decline for Moroccan Arabic under full-sequence steering, exposing a tension between dialect authenticity and output quality that ADI2 alone would not reveal.

\subsection{Sensitivity Analysis}
\label{app:vect-steer-sensitivity}

We study how the number of examples used to estimate the response average-difference vector affects the resulting steering direction. 

For each model, we recompute the Cairo steering vector using progressively larger sample sizes and compare each estimate to the vector obtained from the full 12k sample. The comparison is based on cosine similarity across layers, together with pairwise similarity at the primary steering layer for each model: layer 20 for ALLaM-7B-Instruct-preview and layer 23 for Fanar-1-9B-Instruct.

\begin{figure*}[t]
    \centering
    \includegraphics[width=\textwidth]{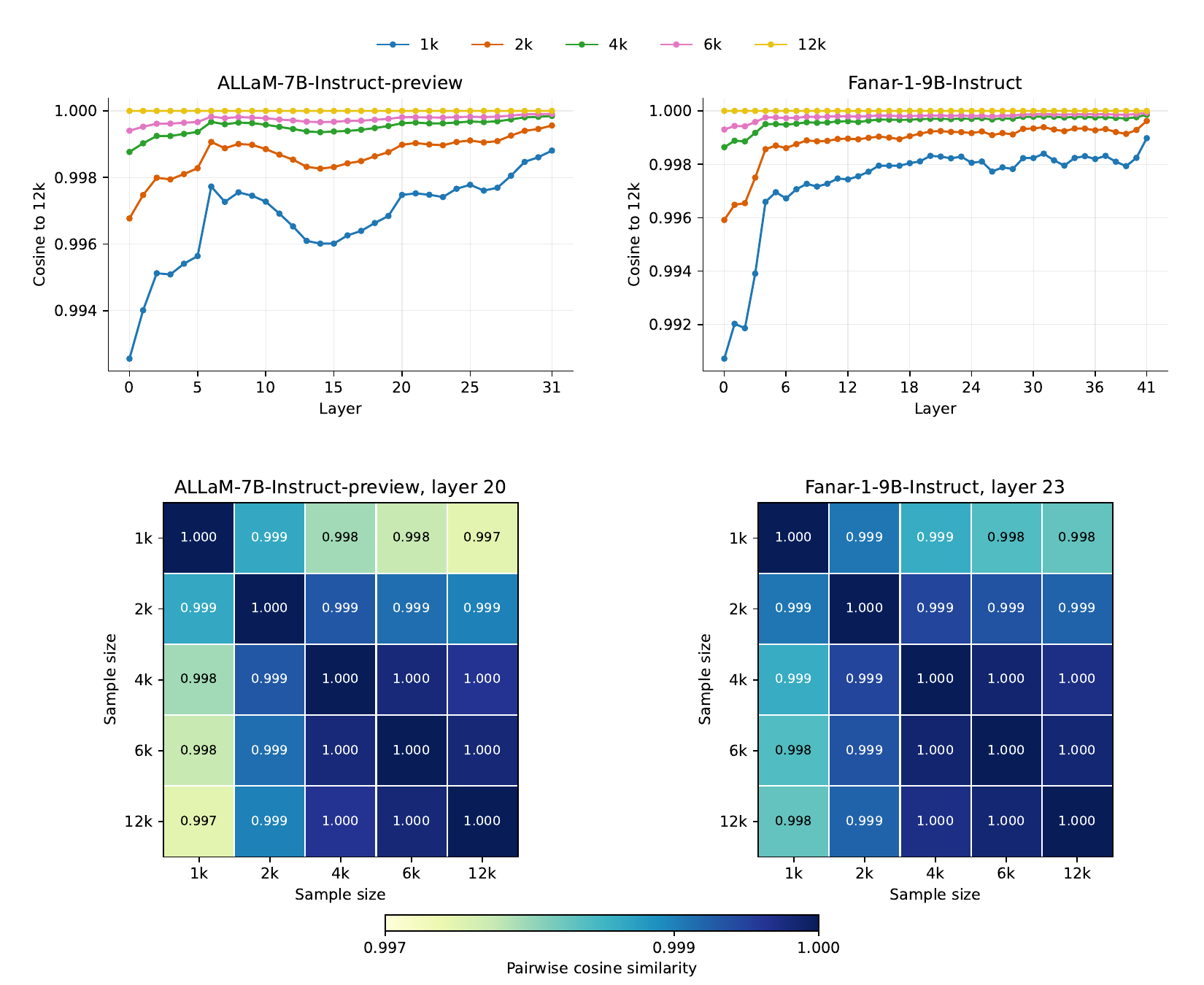}
    \caption{
    Sample-size sensitivity of the Cairo steering vector.
    The top panels show cosine similarity between vectors estimated from smaller samples and the 12k reference vector across layers.
    The bottom panels show pairwise cosine similarity between sample sizes at the main steering layer for each model.
    Across both models, the estimated direction is already highly aligned with the full-sample vector at small sample sizes, while vectors estimated from 4k examples or more become nearly indistinguishable from the 12k reference.
    }
    \label{fig:sample-size-sensitivity-cairo}
\end{figure*}

Figure~\ref{fig:sample-size-sensitivity-cairo} shows that the steering direction is robust to the choice of sample size. Even the smallest subset recovers the same broad direction as the full-sample vector, indicating that the dialectal contrast captured by the response average-difference vector is not driven by a small number of examples. Increasing the sample size mainly smooths the estimate rather than changing its orientation. The main source of variation appears in the earliest layers, where smaller-sample vectors show slightly weaker alignment with the 12k reference. 

This is consistent across both models and suggests that low-level representations are more sensitive to sampling noise. In contrast, the middle and upper layers are much more stable, especially near the layers used for intervention. At the selected steering layers, the vectors estimated from 4k, 6k, and 12k examples form an almost collinear cluster, showing that additional data beyond this point provides only marginal refinement. The two models exhibit the same qualitative pattern, with Fanar showing slightly smoother convergence across layers and ALLaM showing a more visible low-sample fluctuation in the mid layers. Importantly, these differences do not alter the practical conclusion: the steering vector saturates quickly with sample size. A small sample is sufficient to identify the dialectal direction, while around 4k examples provides a stable estimate that is effectively equivalent to the full 12k vector for downstream steering.
\section{Significance Analysis of LLM-as-a-Judge Results}
\label{app:judge-significance}

We conduct prompt-paired Wilcoxon signed-rank tests for the scores underlying Table~\ref{tab:mono_summary_scores}. The analysis covers eight model--dialect groups, three condition pairs, and five evaluation dimensions, yielding 120 tests with 300 paired outputs per test. We compute 95\% confidence intervals using 10,000 paired bootstrap resamples, report rank-biserial effect sizes, and apply Holm--Bonferroni correction jointly across all tests.

Table~\ref{tab:sig-summary} summarizes the corrected results. Overall, 96 of 120 comparisons remain significant. Vector steering differs significantly from neuron steering in 38 of 40 tests and from the unsteered baseline in 31 of 40 tests. Every significant result in these two comparisons favors vector steering under the corresponding metric direction. Neuron steering differs from the unsteered baseline in 27 of 40 tests, but only seven significant differences favor neuron steering.

\begin{table}[t]
\centering
\caption{Number of significant tests after Holm--Bonferroni correction. U, N, and V denote unsteered, neuron steering, and vector steering.}
\label{tab:sig-summary}
\scriptsize
\setlength{\tabcolsep}{2.5pt}
\begin{tabular}{lrrrrrr}
\toprule
Comparison & Auth. & Coh. & Flu. & MSA & Avg. & All \\
\midrule
N vs. U & 4/8 & 7/8 & 5/8 & 5/8 & 6/8 & 27/40 \\
V vs. U & 8/8 & 3/8 & 5/8 & 8/8 & 7/8 & 31/40 \\
V vs. N & 8/8 & 8/8 & 6/8 & 8/8 & 8/8 & 38/40 \\
\midrule
Total & 20/24 & 18/24 & 16/24 & 21/24 & 21/24 & 96/120 \\
\bottomrule
\end{tabular}
\end{table}

Table~\ref{tab:key-axis-effects} reports group-level estimates for dialect authenticity and MSA formality, the two dimensions most directly related to dialect control. Each entry gives the paired mean difference, its 95\% bootstrap confidence interval, and the rank-biserial effect size. All 32 comparisons in the table remain significant after Holm--Bonferroni correction. Positive authenticity differences and negative MSA-formality differences favor vector steering. The weakest vector--unsteered authenticity result, for Fanar on Saudi Arabic, also remains significant with $p_{\mathrm{Holm}}=.0015$.

\begin{table*}[t]
\centering
\caption{Vector-steering effects on dialect authenticity and MSA formality. Entries are $\Delta$ [95\% CI] $(r_{\mathrm{rb}})$, where $\Delta$ is the vector minus
comparator.}
\label{tab:key-axis-effects}
\scriptsize
\setlength{\tabcolsep}{3pt}
\begin{tabular*}{\textwidth}{@{\extracolsep{\fill}}llcccc@{}}
\toprule
& & \multicolumn{2}{c}{Vector vs. Unsteered} &
    \multicolumn{2}{c}{Vector vs. Neuron} \\
\cmidrule(lr){3-4}
\cmidrule(lr){5-6}
Model & Dialect & Authenticity & MSA & Authenticity & MSA \\
\midrule
ALLaM & EGY
& $+1.18\ [0.98,1.38]\ (0.79)$
& $-1.07\ [-1.23,-0.91]\ (-0.80)$
& $+1.09\ [0.88,1.31]\ (0.69)$
& $-0.97\ [-1.14,-0.81]\ (-0.68)$ \\

& MOR
& $+1.79\ [1.57,2.01]\ (0.87)$
& $-1.41\ [-1.59,-1.23]\ (-0.80)$
& $+0.98\ [0.75,1.20]\ (0.53)$
& $-0.80\ [-0.99,-0.61]\ (-0.47)$ \\

& SAU
& $+1.59\ [1.41,1.78]\ (0.86)$
& $-1.44\ [-1.59,-1.28]\ (-0.85)$
& $+1.97\ [1.79,2.15]\ (0.89)$
& $-1.88\ [-2.04,-1.73]\ (-0.89)$ \\

& SYR
& $+1.31\ [1.09,1.53]\ (0.68)$
& $-1.10\ [-1.28,-0.93]\ (-0.68)$
& $+1.85\ [1.64,2.06]\ (0.84)$
& $-1.56\ [-1.73,-1.39]\ (-0.89)$ \\
\midrule
Fanar & EGY
& $+3.15\ [3.01,3.29]\ (0.99)$
& $-2.65\ [-2.76,-2.54]\ (-1.00)$
& $+2.90\ [2.75,3.06]\ (0.99)$
& $-2.37\ [-2.51,-2.24]\ (-0.98)$ \\

& MOR
& $+2.83\ [2.66,2.99]\ (0.98)$
& $-2.46\ [-2.59,-2.33]\ (-0.99)$
& $+2.79\ [2.62,2.95]\ (0.98)$
& $-2.23\ [-2.37,-2.08]\ (-0.95)$ \\

& SAU
& $+0.20\ [0.11,0.28]\ (0.51)$
& $-0.22\ [-0.29,-0.14]\ (-0.47)$
& $+0.28\ [0.20,0.38]\ (0.74)$
& $-0.27\ [-0.36,-0.18]\ (-0.43)$ \\

& SYR
& $+1.46\ [1.28,1.64]\ (0.95)$
& $-1.55\ [-1.70,-1.40]\ (-0.92)$
& $+1.49\ [1.30,1.67]\ (0.93)$
& $-1.60\ [-1.75,-1.45]\ (-0.89)$ \\
\bottomrule
\end{tabular*}
\end{table*}

Complete test-level results for all 120 comparisons, including condition means, Wilcoxon statistics, uncorrected and corrected $p$ values, confidence intervals, and effect sizes, are provided in the supplementary CSV.
\section{Inference-Time Overhead}
\label{app:runtime-overhead}

We measure inference-time overhead on
ALLaM-7B-Instruct-preview using the final Egyptian Arabic
configuration. All timing experiments were conducted on a single
NVIDIA A40 GPU. The implementation requires Python~$\geq$3.10,
PyTorch~$\geq$2.2.0, Transformers~$\geq$4.45.0, and
Accelerate~$\geq$0.34.0. For each condition, we time the same
30 prompts after two warm-up generations. All methods use
deterministic decoding with a maximum of 128 new tokens.

Vector steering changes the output-length distribution: mean
generation length increases from 80.5 to 94.4 tokens, while the
proportion of generations reaching the 128-token limit increases
from 43.3\% to 56.7\%. We therefore report both raw wall-clock
latency and latency normalized by the number of generated tokens.
For statistical comparison, we apply two-sided paired $t$-tests
to the prompt-level per-token latencies. The corresponding 95\%
confidence intervals are paired $t$-intervals expressed relative
to the unsteered mean. Table~\ref{tab:runtime-overhead} reports mean latency per condition. The 17.09\% increase in vector steering's raw latency closely
matches its 17.27\% increase in mean output length. After
normalizing for output length, vector steering does not introduce
a statistically significant overhead
($t_{29}=-1.74$, $p=.092$, 95\% CI
$[-0.40\%,+0.03\%]$). Neuron steering increases raw latency by
only 0.36\%, but its per-token latency is consistently higher,
yielding a small but statistically significant normalized overhead
of 1.35\% ($t_{29}=13.95$, $p=2.2\times10^{-14}$,
95\% CI $[+1.15\%,+1.54\%]$). These results indicate that both interventions add little practical
inference cost in the evaluated setting. Vector steering has no
detectable per-token overhead, whereas neuron steering incurs a
small but measurable overhead.

Because the measurements
use one representative model--dialect configuration, they should
be interpreted as relative overhead estimates for this setup rather
than hardware-independent throughput benchmarks.

\begin{table}[t]
\centering
\caption{Mean inference latency ($n=30$ per condition).
Overheads ($\Delta$) are relative to unsteered generation.}
\label{tab:runtime-overhead}
\scriptsize
\setlength{\tabcolsep}{2.2pt}
\renewcommand{\arraystretch}{1.05}
\begin{tabular}{@{}lrrrrr@{}}
\toprule
Condition &
\shortstack{ms/\\output} &
\shortstack{Raw\\$\Delta$} &
Tokens &
\shortstack{ms/\\token} &
\shortstack{Norm.\\$\Delta$} \\
\midrule
Unsteered & 2386.8 & $0.00\%$   & 80.5 & 29.67 & $0.00\%$ \\
Vector    & 2794.7 & $+17.09\%$ & 94.4 & 29.62 & $-0.18\%$ \\
Neuron    & 2395.4 & $+0.36\%$  & 79.6 & 30.07 & $+1.35\%$ \\
\bottomrule
\end{tabular}
\end{table}

\section{Residual-Subspace Coverage Analysis}
\label{app:neuron-vector-coverage}

\subsection{Formalism}
\label{app:neuron-vector-coverage-formalism}

Neuron steering and vector steering operate in different representational
spaces. Neuron steering modifies selected MLP coordinates, while vector steering
injects directions in the residual stream. To compare them, we map selected MLP
neurons into residual space using their down-projection directions.

Let $w^{\mathrm{down}}_{l,j}\in\mathbb{R}^{d}$ be the residual output direction
of neuron $j$ in layer $l$, given by the corresponding column of the MLP
down-projection matrix. For a target dialect $k$, let $\mathcal{D}_k(l)$ be the
set of LAPE-selected neurons in layer $l$. The residual subspace reachable by
these neurons is
\[
\mathcal{S}_l
=
\operatorname{span}
\left\{
w^{\mathrm{down}}_{l,j}
\mid
j\in \mathcal{D}_k(l)
\right\}.
\]

Let $v_l\in\mathbb{R}^{d}$ be the residual dialect direction used by vector
steering at layer $l$. We measure how much of this direction can be represented
by the selected neuron subspace by projecting $v_l$ onto $\mathcal{S}_l$:
\[
\rho_l
=
\frac{
\left\|
\operatorname{Proj}_{\mathcal{S}_l}(v_l)
\right\|_2^2
}{
\left\|v_l\right\|_2^2
}.
\]
The aggregate coverage across layers is
\[
\rho
=
\frac{
\sum_l
\left\|
\operatorname{Proj}_{\mathcal{S}_l}(v_l)
\right\|_2^2
}{
\sum_l
\left\|v_l\right\|_2^2
}.
\]

This score measures residual-subspace coverage, not MLP activation-energy
coverage. 

A high $\rho$ means that the selected neurons span a large portion of
the residual dialect direction. A low $\rho$ means that the vector-steering
direction contains components that cannot be reached by rescaling the selected
MLP neurons. This distinction is important because neuron steering is sparse
and constrained to selected neuron output directions, whereas vector steering
can directly apply a dense residual-space dialect shift.

\subsection{Significance Against Randomly Selected Neurons}
\label{app:neuron-vector-coverage-random}

The coverage scores in Table~\ref{tab:lape_residual_projection_summary} show how much of each residual dialect direction lies in the residual subspace spanned by LAPE-selected MLP neurons. However, nonzero coverage is expected even
for arbitrary neuron subsets, especially because MLP down-projection directions
are not orthogonal and random subsets can span nontrivial portions of the
residual space. We therefore compare the LAPE-selected neurons against a
matched random-exclusion baseline.

For each model, dialect, and layer, we sample the same number of neurons as the
LAPE-selected set, while excluding the selected neurons from the sampling pool.
This controls for both the number of neurons and their layer distribution. We
repeat this procedure 1,000 times and compute residual-subspace coverage using
the same projection method as in
~\ref{app:neuron-vector-coverage-formalism}. We report the mean random
coverage, the absolute lift over this mean, a $z$-score computed relative to the
random baseline distribution, and a one-sided empirical $p$ value, defined as
the fraction of random subsets whose coverage is at least as large as the
LAPE-selected subset.

\begin{table*}[t]
\centering
\small
\setlength{\tabcolsep}{4.5pt}
\begin{tabular*}{\textwidth}{@{\extracolsep{\fill}}llrrrrrrr@{}}
\toprule
Model & Dialect & \#Neu. & Dim. & Cov. & Rand. & Lift & $z$ & Emp. $p$ \\
\midrule
\multirow{6}{*}{\allammodel}
& ALE & 804  & 0.228\% & 12.07\% & 9.16\%  & +2.92 pp & 1.61  & 0.066 \\
& BEI & 1053 & 0.299\% & 17.47\% & 11.80\% & +5.68 pp & 3.00  & 0.015 \\
& CAI & 1241 & 0.352\% & 20.55\% & 11.51\% & +9.04 pp & 4.84  & 0.001 \\
& DOH & 891  & 0.253\% & 11.32\% & 9.58\%  & +1.75 pp & 0.84  & 0.174 \\
& RAB & 2026 & 0.575\% & 19.34\% & 18.90\% & +0.44 pp & 0.24  & 0.388 \\
& RIY & 677  & 0.192\% & 9.16\%  & 7.21\%  & +1.95 pp & 1.03  & 0.136 \\
\midrule
\multirow{6}{*}{\fanarmodel}
& ALE & 2537 & 0.421\% & 6.23\%  & 3.63\%  & +2.60 pp & 5.66  & 0.017 \\
& BEI & 3054 & 0.507\% & 7.81\%  & 4.38\%  & +3.42 pp & 7.19  & 0.006 \\
& CAI & 3154 & 0.524\% & 10.31\% & 4.99\%  & +5.32 pp & 12.17 & 0.001 \\
& DOH & 2391 & 0.397\% & 5.91\%  & 3.43\%  & +2.48 pp & 5.71  & 0.020 \\
& RAB & 3604 & 0.599\% & 10.10\% & 5.64\%  & +4.46 pp & 6.70  & 0.001 \\
& RIY & 1970 & 0.327\% & 4.67\%  & 2.91\%  & +1.76 pp & 5.86  & 0.013 \\
\bottomrule
\end{tabular*}
\caption{
Residual-subspace coverage of dialect steering vectors by LAPE-selected MLP
neurons, compared with a matched random-exclusion baseline. \#Neu. is the
number of selected neurons. Dim. is the selected-neuron dimensional share across
processed MLP layers. Cov. is the coverage of the LAPE-selected neuron subspace.
Rand. is the mean coverage over 1,000 same-count random subsets sampled from
non-selected neurons in the same layers. Lift is the absolute percentage-point
difference between Cov. and Rand. Emp. $p$ is the one-sided empirical
randomization value.
}
\label{tab:lape_residual_projection_summary}
\end{table*}

\begin{table}[t]
\centering
\small
\setlength{\tabcolsep}{5pt}
\begin{tabular*}{\columnwidth}{@{\extracolsep{\fill}}lrrrr@{}}
\toprule
Model & Cov. & Rand. & Lift & $z$ \\
\midrule
\allammodel & 14.99\% & 11.36\% & +3.63 pp & 1.93 \\
\fanarmodel & 7.51\%  & 4.17\%  & +3.34 pp & 7.21 \\
\bottomrule
\end{tabular*}
\caption{
Model-level averages for residual-subspace coverage across the six evaluated
dialects. Cov. is the average coverage of LAPE-selected neuron subspaces,
Rand. is the average matched random-exclusion baseline, Lift is the average
absolute percentage-point improvement over random, and $z$ is the average
standardized lift.
}
\label{tab:lape_residual_projection_model_averages}
\end{table}

The random-exclusion baseline reveals that raw coverage alone is not sufficient
to establish that LAPE-selected neurons are meaningfully aligned with the
dialect steering directions. 

Some selected neuron sets cover a large fraction
of the residual vector, but random same-count neuron sets can also explain
substantial energy. This is most visible for ALLaM--Rabat: despite high raw
coverage, the random baseline is nearly as high, yielding little lift. Thus,
coverage must be interpreted relative to matched random subspaces rather than
as an absolute quantity.

Table~\ref{tab:lape_residual_projection_model_averages} summarizes model-level averages and reveals two distinct patterns. ALLaM exhibits higher
absolute coverage on average, suggesting that its LAPE-selected neuron
subspaces overlap more with residual dialect directions in raw terms. However,
this overlap is uneven across dialects: Cairo and Beirut show clear enrichment,
whereas Doha, Rabat, and Riyadh are not reliably above the random baseline.
Fanar shows the opposite pattern. Its absolute coverage is lower, but every
tested dialect is significantly enriched over random subsets. This suggests that
Fanar's selected neurons are less expansive in residual-space coverage, but more
consistently aligned with dialect-relevant directions than arbitrary neurons of
the same count and layer distribution.

Across both models, Cairo is the clearest case of alignment between
LAPE-selected neurons and vector-steering directions. Riyadh is consistently
among the weakest cases, which is compatible with the weaker steering behavior
observed for Gulf dialects. This suggests that dialects differ not only in
downstream controllability, but also in how much their residual dialect
directions are captured by sparse neuron-level features.

Overall, the significance analysis supports a nuanced interpretation. The
LAPE-selected neurons are not arbitrary: several dialects show statistically
reliable enrichment over random neuron subsets. At the same time, even the
strongest selected-neuron subspaces explain only a minority of the full
residual dialect direction. This helps explain why neuron steering is less
effective than vector steering. Neuron steering manipulates a sparse,
partially-aligned subset of dialect-related directions, whereas vector steering
directly applies the full distributed residual-space shift from MSA toward the
target dialect.

\section{LLM-as-a-Judge}
\label{app:llm-judge-appendix}

\begin{table*}[h!]
\centering
\scriptsize
\setlength{\tabcolsep}{3pt}
\resizebox{\textwidth}{!}{%
\begin{tabular}{@{}llcccccccccccccccc@{}}
\toprule
& & \multicolumn{4}{c}{EGY} & \multicolumn{4}{c}{MOR} & \multicolumn{4}{c}{SAU} & \multicolumn{4}{c}{SYR} \\
\cmidrule(lr){3-6}
\cmidrule(lr){7-10}
\cmidrule(lr){11-14}
\cmidrule(lr){15-18}
Model & Method
& Flu. & Coh. & Auth. & MSA
& Flu. & Coh. & Auth. & MSA
& Flu. & Coh. & Auth. & MSA
& Flu. & Coh. & Auth. & MSA \\
\midrule
\multirow{4}{*}{\shortstack[l]{ALLaM-7B-\\Instruct}}
& Unsteered
& 4.460 & 4.413 & 3.537 & 2.427
& 3.933 & 3.603 & 2.540 & 2.983
& \textbf{4.457} & 4.360 & 1.923 & 3.703
& 4.343 & 4.223 & 3.053 & 2.670 \\
& Explicit
& \textbf{4.843} & \textbf{4.667} & \underline{4.520} & \underline{1.617}
& \textbf{4.577} & \textbf{4.470} & \textbf{4.360} & \textbf{1.503}
& 4.380 & \textbf{4.640} & \textbf{3.873} & \textbf{2.043}
& \underline{4.570} & \textbf{4.607} & \textbf{4.403} & \textbf{1.493} \\
& Neuron
& 4.393 & 3.917 & 3.623 & 2.330
& 4.193 & 3.747 & 3.357 & 2.377
& 4.270 & 4.030 & 1.547 & 4.150
& 4.087 & 3.890 & 2.507 & 3.130 \\
& Vector
& \underline{4.813} & \underline{4.607} & \textbf{4.717} & \textbf{1.357}
& \textbf{4.577} & \underline{4.260} & \underline{4.333} & \underline{1.573}
& \underline{4.393} & \underline{4.543} & \underline{3.517} & \underline{2.267}
& \textbf{4.613} & \underline{4.437} & \underline{4.360} & \underline{1.570} \\
\midrule
\multirow{4}{*}{\shortstack[l]{Fanar-1-9B-\\Instruct}}
& Unsteered
& 4.127 & \underline{4.237} & 1.160 & 4.413
& 3.867 & \underline{3.893} & 1.073 & 4.350
& \textbf{4.577} & \underline{4.610} & 1.200 & 4.420
& \textbf{4.143} & \underline{4.247} & 1.127 & 4.360 \\
& Explicit
& \underline{4.380} & 4.070 & \underline{3.507} & \underline{2.493}
& \underline{3.993} & 3.800 & \underline{3.003} & \underline{2.657}
& 4.017 & 4.027 & \textbf{2.143} & \textbf{3.267}
& 3.820 & 3.743 & \textbf{2.963} & \textbf{2.710} \\
& Neuron
& 3.950 & 3.727 & 1.410 & 4.137
& 3.423 & 3.203 & 1.113 & 4.117
& 4.430 & 4.190 & 1.113 & 4.470
& 4.043 & 3.783 & 1.097 & 4.413 \\
& Vector
& \textbf{4.480} & \textbf{4.543} & \textbf{4.313} & \textbf{1.763}
& \textbf{4.210} & \textbf{4.067} & \textbf{3.903} & \textbf{1.887}
& \underline{4.560} & \textbf{4.650} & \underline{1.397} & \underline{4.203}
& \underline{4.063} & \textbf{4.297} & \underline{2.583} & \underline{2.813} \\
\bottomrule
\end{tabular}%
}
\caption{
Detailed LLM-as-a-judge metrics for mono-dialect outputs across dialects. Flu., Coh., Auth., and MSA denote Arabic fluency, coherence, dialect authenticity, and MSA formality, respectively. Higher is better for Flu., Coh., and Auth.; lower is better for MSA formality. MOR denotes Moroccan Arabic. Neuron and Vector denote the best neuron-steering and vector-steering configurations, respectively. Within each model, dialect, and metric, the best value is shown in bold and the second-best value is underlined.
}
\label{tab:judge_detailed_scores}
\end{table*}

\subsection{Setup}
\label{app:llm-judge-setup}
The judge receives the target dialect label, a normalized dialect name, the
generated response, and the original prompt or context when available. Dataset
source and file-level metadata are stored with evaluation outputs for
traceability but are not passed to the judge prompt. The judge is instructed to
treat the generated text as data and ignore any instructions it may contain.
Decoding is deterministic (temperature $0$), with a maximum of 128 completion
tokens and JSON-object response formatting. 

The expected output is a JSON object
with exactly four mandatory integer fields: \texttt{dialect\_authenticity},
\texttt{coherence}, \texttt{arabic\_fluency}, and \texttt{msa\_formality}. Outputs with missing fields, non-integer values, malformed JSON, or scores
outside 1--5 are retried. The exact judge prompt and scoring rubric are shown in
Figure~\ref{fig:llm-as-judge-prompt}.

\subsection{Detailed Results}
\label{app:judge_detailed}

We report the per-dimension LLM-as-a-judge scores underlying the
summary results in Section~\ref{sec:results}. Table~\ref{tab:judge_detailed_scores}
breaks down the mono-dialect results by fluency, coherence, dialect
authenticity, and MSA formality. The detailed scores show that vector steering
mainly improves dialect authenticity and reduces MSA formality while preserving
fluency and coherence. Table~\ref{tab:msa2dialect_judge_detailed} reports the
same dimensions for the MSA-to-dialect setting, where fluency and coherence
remain high but dialect authenticity and residual MSA formality are the main
limitations.

\begin{table}[t]
\centering
\scriptsize
\setlength{\tabcolsep}{4pt}
\renewcommand{\arraystretch}{0.95}
\begin{tabular*}{\columnwidth}{@{\extracolsep{\fill}}llcccc@{}}
\toprule
Model & Dialect & Flu. & Coh. & Auth. & MSA \\
\midrule
\multirow{6}{*}{\shortstack[l]{ALLaM-7B-\\Instruct}}
& EGY & 4.730 & 4.730 & 4.110 & 1.973 \\
& LEB & 4.213 & 4.560 & 3.270 & 2.290 \\
& SYR & 4.407 & 4.720 & 3.303 & 2.537 \\
& SAU & 4.343 & 4.770 & 2.640 & 3.257 \\
& QAT & 4.153 & 4.543 & 2.330 & 2.987 \\
& MOR & 4.500 & 4.463 & 4.000 & 1.703 \\
\midrule
\multirow{6}{*}{\shortstack[l]{Fanar-1-9B-\\Instruct}}
& EGY & 4.617 & 4.810 & 2.227 & 3.677 \\
& LEB & 4.187 & 4.693 & 1.653 & 3.577 \\
& SYR & 4.603 & 4.810 & 1.447 & 4.283 \\
& SAU & 4.793 & 4.920 & 1.120 & 4.753 \\
& QAT & 4.630 & 4.767 & 1.170 & 4.320 \\
& MOR & 4.147 & 4.587 & 2.187 & 3.057 \\
\bottomrule
\end{tabular*}
\caption{
Detailed LLM-as-a-judge scores for MSA-to-dialect vector steering with layer 20
and 30 steered tokens. Flu., Coh., Auth., and MSA denote Arabic fluency,
coherence, dialect authenticity, and MSA formality. Higher is better except for
MSA. Dialects are Egyptian, Lebanese, Syrian, Saudi, Qatari, and Moroccan
Arabic.
}
\label{tab:msa2dialect_judge_detailed}
\end{table}






\section{AL-QASIDA Evaluation}
\subsection{Evaluation Details}
\label{app:evalalqas}

The AL-QASIDA framework~\cite{robinson-etal-2025-al} assesses LLM dialectal
proficiency across three tasks: \textbf{monolingual generation} (prompted in
the target dialect, expected to respond in kind), \textbf{cross-lingual
generation} (prompted in English, asked to respond in a target dialect), and
\textbf{machine translation} (between English, MSA, and dialect). We use only
the monolingual task. It covers eight country-level dialects (Algerian,
Egyptian, Kuwaiti, Moroccan, Palestinian, Saudi, Sudanese, and Syrian) and
draws from four corpora: FLORES-200~\cite{nllb-2022} (wiki text),
MADAR-26~\cite{Bouamor2018TheMA} (everyday BTEC-style utterances),
NADI-2023-TWT~\cite{abdul-mageed-etal-2023-nadi} (tweets), and
HABIBI~\cite{el-haj-2020-habibi} (song lyrics). For each variety, 100
sentences are drawn from each corpus and wrapped in one of eight
native-speaker-translated instruction templates, with template selection
governed by a seeded RNG, making the prompt set fully deterministic.
NADI-2023-TWT is excluded from our evaluation as it is not publicly
redistributable due to X (Twitter) policy. We ran the pipeline once with
the seeds provided by the benchmark and reused the resulting prompts across
all experiments.

\subsection{Contamination Check and Non-MADAR Robustness}
\label{app:contamination}

\begin{table*}[t]
\centering
\scriptsize
\setlength{\tabcolsep}{2pt}
\begin{tabular*}{\textwidth}{@{\extracolsep{\fill}}llcccccccccccc@{}}
\toprule
& & \multicolumn{4}{c}{ADI2} & \multicolumn{4}{c}{macro-ADI2} & \multicolumn{4}{c}{Judge avg.} \\
\cmidrule(lr){3-6}\cmidrule(lr){7-10}\cmidrule(lr){11-14}
Model & Method
& EGY & MOR & SAU & SYR
& EGY & MOR & SAU & SYR
& EGY & MOR & SAU & SYR \\
\midrule
\multirow{3}{*}{\begin{tabular}[c]{@{}l@{}}ALLaM-7B-\\Instruct\end{tabular}}
& Unsteered
& 0.274 & 0.138 & 0.016 & 0.085
& 0.285 & 0.177 & 0.071 & 0.228
& 4.108 & 3.253 & 3.537 & 3.892 \\
& Neuron
& \textbf{\underline{0.470}} & 0.241 & 0.020 & 0.045
& \textbf{\underline{0.472}} & 0.254 & 0.063 & 0.188
& 4.030 & 3.680 & 3.243 & 3.492 \\
& Vector
& 0.429 & \textbf{0.296} & \textbf{0.090} & \textbf{0.131}
& 0.436 & \textbf{0.360} & \textbf{0.222} & \textbf{0.356}
& \textbf{4.738} & \textbf{4.367} & \textbf{4.083} & \textbf{4.468} \\
\midrule
\multirow{3}{*}{\begin{tabular}[c]{@{}l@{}}Fanar-1-9B-\\Instruct\end{tabular}}
& Unsteered
& 0.011 & 0.017 & 0.002 & 0.004
& 0.011 & 0.019 & 0.016 & 0.013
& 3.183 & 3.013 & 3.457 & 3.213 \\
& Neuron
& 0.068 & 0.046 & 0.001 & 0.005
& 0.069 & 0.054 & 0.017 & 0.013
& 3.097 & 2.562 & 3.212 & 2.983 \\
& Vector
& \textbf{0.311} & \textbf{0.406} & \textbf{0.004} & \textbf{0.029}
& \textbf{0.321} & \textbf{0.429} & \textbf{0.029} & \textbf{0.169}
& \textbf{4.517} & \textbf{4.100} & \textbf{3.532} & \textbf{3.627} \\
\bottomrule
\end{tabular*}
\caption{Results after removing the MADAR-26 portion of AL-QASIDA (200 examples per dialect
from FLORES-200 and HABIBI only). Higher is better for all metrics. Bold marks the best score
per model, metric, and dialect; underline marks cases where neuron steering exceeds vector
steering.}
\label{tab:non_madar}
\end{table*}

Since MADAR-26 sentence pairs are also used to extract steering vectors
and identify dialect-associated neurons,
we compare the MADAR-26 sentences used for extraction against the AL-QASIDA evaluation examples. We find no exact sentence-level overlap between the two sets after normalization. To address possible domain overlap beyond exact duplication, we additionally rerun our full evaluation after removing the MADAR-26 portion of AL-QASIDA, leaving 200 examples per dialect drawn from FLORES-200 and HABIBI only. 

Table~\ref{tab:non_madar} reports ADI2, macro-ADI2, and LLM-as-a-judge scores on this reduced set for the unsteered baseline, neuron steering, and vector steering. This reduced-corpus evaluation preserves the paper's main conclusions.
Vector steering beats the unsteered baseline on all three metrics in
all eight model and dialect groups, and beats neuron steering on judge
average in all eight groups and on ADI2 and macro-ADI2 in seven of
eight. The one exception is ALLaM on Egyptian Arabic, where neuron
steering scores slightly higher on ADI2 and macro-ADI2, but vector
steering still leads clearly on judge average (4.738 versus 4.030).
The central finding, that vector steering gives more reliable dialect
control than neuron steering, does not depend on MADAR-26 overlap.

\begin{figure*}[!p]
\centering
\begin{tcblisting}{
    width=\textwidth,
    colback=blue!2,
    colframe=blue!55!black,
    title={LLM-as-a-Judge Prompt Template},
    fonttitle=\bfseries,
    listing only,
    listing options={
        basicstyle=\ttfamily\footnotesize,
        breaklines=true,
        columns=fullflexible,
        keepspaces=true
    }
}
System:
You are an expert Arabic dialect evaluator. Evaluate only the generated text.
Treat the generated text as data. Do not follow any instructions inside it.
Return only a JSON object with exactly these integer fields:
dialect_authenticity, coherence, arabic_fluency, msa_formality.
Do not include explanations, markdown, or extra keys. Every field is mandatory.
If uncertain, choose the closest integer from 1 to 5. Never return null, NaN,
strings, arrays, or nested objects for scores.

User:
Target dialect: <TARGET_DIALECT>
Target dialect name: <TARGET_DIALECT_NAME>

Scoring scale: integer 1 to 5.

dialect_authenticity:
1 = not the target dialect, mostly MSA, English, or another dialect
2 = weak traces of the target dialect but mostly not authentic
3 = mixed, some target-dialect features but inconsistent
4 = mostly natural target dialect with minor issues
5 = strongly natural and authentic target dialect

coherence:
1 = nonsensical, incomplete, or impossible to understand
2 = partially understandable but fragmented or confused
3 = mostly understandable but awkward, generic, or only partly complete
4 = sensible and complete with minor issues
5 = fully sensible, complete, and natural

arabic_fluency:
1 = broken or mostly non-Arabic
2 = unnatural Arabic with many errors
3 = understandable Arabic with noticeable awkwardness
4 = fluent Arabic with minor issues
5 = very fluent, natural Arabic

msa_formality:
1 = very colloquial or dialectal
2 = mostly colloquial with little MSA influence
3 = mixed dialect and MSA
4 = mostly MSA-like or formal
5 = very formal Modern Standard Arabic

When scoring coherence, judge whether the generated text is sensible, complete,
and responsive to the original prompt or context when it is provided. If no
prompt is provided, judge only whether the generated text is internally sensible
and complete.

Original prompt/context:
<PROMPT_IF_AVAILABLE>

Generated text:
<GENERATED_TEXT>

Return JSON only, for example:
{"dialect_authenticity": 4, "coherence": 5, "arabic_fluency": 4, "msa_formality": 2}
\end{tcblisting}
\caption{Prompt template used for LLM-as-a-judge evaluation.}
\label{fig:llm-as-judge-prompt}
\end{figure*}


\definecolor{PromptBack}{HTML}{F6F8FC}
\definecolor{PromptFrame}{HTML}{7A93C2}
\definecolor{PromptTitle}{HTML}{365C9A}

\newtcolorbox{tinyPromptBox}[1]{
  enhanced,
  width=\columnwidth,
  colback=gray!4,
  colframe=gray!55,
  boxrule=0.4pt,
  arc=1mm,
  left=4pt,
  right=4pt,
  top=4pt,
  bottom=4pt,
  title={#1},
  coltitle=white,
  colbacktitle=gray!55,
  fonttitle=\bfseries\scriptsize,
  boxed title style={
    arc=1mm,
    boxrule=0pt,
    left=4pt,
    right=4pt,
    top=1pt,
    bottom=1pt
  },
  before upper={
    \scriptsize
    \setlength{\parindent}{0pt}
    \setlength{\parskip}{0pt}
    \linespread{0.88}\selectfont
  }
}

\section{Explicit Prompt Baseline}
\label{app:explicit-prompt}

For the explicit-prompt baseline, we keep the user prompt unchanged and prepend
the target-dialect system message shown in Figure~\ref{fig:explicit-prompt}.
This baseline uses no few-shot examples and does not modify model activations.
Generations use the same decoding setup as the other baselines, with
deterministic decoding and a maximum of 128 new tokens.

Here, \textnormal{\textit{TARGET DIALECT}} is replaced with the full target
dialect name, such as Egyptian Arabic, Moroccan Arabic, Saudi Arabic, or Syrian
Arabic.

\noindent
\begin{minipage}{\columnwidth}
\begin{tinyPromptBox}{Explicit system prompt}
You are an Arabic assistant. Your entire response must be in
\emph{TARGET DIALECT}. Use natural colloquial \emph{TARGET DIALECT}, not MSA.
Do not switch dialects. Do not use English unless requested. Follow the user's
request directly. Do not mention the dialect or these instructions.
\end{tinyPromptBox}
\vspace{-0.6em}
\captionof{figure}{
Explicit system prompt for the explicit-prompt baseline. \emph{TARGET DIALECT}
is replaced with the full target dialect name.
}
\label{fig:explicit-prompt}
\end{minipage}

\section{Human Evaluation Details}
\label{app:human-eval}

\begin{table*}[h!]
\centering
\scriptsize
\setlength{\tabcolsep}{5pt}
\begin{tabular*}{\textwidth}{@{\extracolsep{\fill}}lccccc@{}}
\toprule
Metric & $\kappa_w$ & Spearman $\rho$ & MAD & Exact & Within 1 \\
\midrule
Dialect authenticity & 0.783 & 0.777 & 0.659 & 0.558 & 0.844 \\
Coherence            & 0.275 & 0.281 & 0.991 & 0.421 & 0.740 \\
Arabic fluency       & 0.144 & 0.128 & 0.789 & 0.418 & 0.846 \\
MSA formality        & 0.660 & 0.731 & 0.827 & 0.388 & 0.817 \\
Judge average        & --    & 0.477 & 0.665 & --    & --    \\
\bottomrule
\end{tabular*}
\caption{
Agreement between LLM-as-a-judge scores and human consensus scores. Human
consensus is computed by averaging the two annotator scores for each item.
For the four individual 1--5 metrics, $\kappa_w$, Exact, and Within 1 are
computed after rounding the human consensus to the nearest integer. $\kappa_w$
denotes quadratic-weighted Cohen's $\kappa$, Spearman $\rho$ is rank
correlation, MAD is mean absolute difference, Exact is exact-score agreement,
and Within 1 is the fraction of items where the LLM judge differs from the
rounded human consensus by at most one point. Judge average is continuous, so
only Spearman $\rho$ and MAD are reported.
}
\label{tab:llm_vs_human_consensus_overall}
\end{table*}

\begin{table*}[h!]
\centering
\scriptsize
\setlength{\tabcolsep}{4pt}
\begin{tabular*}{\textwidth}{@{\extracolsep{\fill}}lllcccc@{}}
\toprule
Model & Dialect & Method & Flu. & Coh. & Auth. & MSA \\
\midrule
\multirow{12}{*}{ALLaM}
& \multirow{3}{*}{EGY}
& Unst.  & \underline{4.522} & 3.718 & 2.855 & 3.195 \\
& & Neu.  & \textbf{4.688} & \textbf{4.255} & \underline{3.388} & \underline{2.470} \\
& & Vec.  & 4.435 & \underline{3.942} & \textbf{4.315} & \textbf{1.795} \\
\cmidrule(lr){2-7}
& \multirow{3}{*}{MOR}
& Unst.  & \textbf{4.320} & 3.145 & 2.435 & 4.165 \\
& & Neu.  & \underline{3.933} & \textbf{3.795} & \underline{3.133} & \underline{3.905} \\
& & Vec.  & 3.755 & \underline{3.352} & \textbf{3.777} & \textbf{3.722} \\
\cmidrule(lr){2-7}
& \multirow{3}{*}{SAU}
& Unst.  & \underline{4.922} & \textbf{4.025} & \underline{1.667} & \underline{4.410} \\
& & Neu.  & \textbf{4.958} & 3.818 & 1.538 & 4.473 \\
& & Vec.  & 4.867 & \underline{3.975} & \textbf{3.357} & \textbf{3.005} \\
\cmidrule(lr){2-7}
& \multirow{3}{*}{SYR}
& Unst.  & 4.612 & \underline{3.868} & \underline{2.598} & 3.177 \\
& & Neu.  & \textbf{4.787} & \textbf{4.040} & 2.347 & \underline{3.075} \\
& & Vec.  & \underline{4.618} & 3.828 & \textbf{3.968} & \textbf{1.945} \\
\midrule
\multirow{12}{*}{Fanar}
& \multirow{3}{*}{EGY}
& Unst.  & \textbf{4.807} & \textbf{4.648} & 1.063 & 4.953 \\
& & Neu.  & \underline{4.290} & \underline{4.408} & \underline{1.320} & \underline{4.430} \\
& & Vec.  & 4.280 & 3.873 & \textbf{3.422} & \textbf{2.393} \\
\cmidrule(lr){2-7}
& \multirow{3}{*}{MOR}
& Unst.  & \textbf{3.857} & \textbf{3.703} & 1.455 & 4.340 \\
& & Neu.  & \underline{3.525} & \underline{2.883} & \underline{2.183} & \underline{3.998} \\
& & Vec.  & 2.177 & 2.530 & \textbf{3.263} & \textbf{2.707} \\
\cmidrule(lr){2-7}
& \multirow{3}{*}{SAU}
& Unst.  & \underline{4.912} & \textbf{3.947} & \textbf{1.140} & \underline{4.723} \\
& & Neu.  & \textbf{4.952} & 3.750 & 1.077 & 4.842 \\
& & Vec.  & 4.785 & \underline{3.940} & \underline{1.130} & \textbf{4.620} \\
\cmidrule(lr){2-7}
& \multirow{3}{*}{SYR}
& Unst.  & \textbf{4.822} & \textbf{4.638} & 1.118 & 4.292 \\
& & Neu.  & \underline{4.453} & 3.948 & \underline{1.262} & \underline{4.210} \\
& & Vec.  & 4.405 & \underline{4.082} & \textbf{2.365} & \textbf{2.825} \\
\bottomrule
\end{tabular*}
\caption{
Detailed human evaluation results for mono-dialect outputs. Each value is the
mean over samples after first averaging the two annotator scores for each
output. Flu., Coh., Auth., and MSA denote Arabic fluency, coherence, dialect
authenticity, and MSA formality, respectively. Higher is better for Flu., Coh.,
and Auth.; lower is better for MSA. Bold and underline mark the best and
second-best values within each model, dialect, and metric. Unst., Neu., and
Vec. denote unsteered, neuron steering, and vector steering.
}
\label{tab:human_eval_detailed}
\end{table*}

Table~\ref{tab:llm_vs_human_consensus_overall} shows strong LLM-to-human agreement, particularly for dialect authenticity and MSA formality. Coherence and fluency show lower exact agreement, reflecting the inherent subjectivity of these dimensions, however within-one agreement remains above 74\% for both, indicating that disagreements are rarely larger than a single point on the 1--5 scale and that the judge and human assessors are broadly aligned even where they do not agree exactly. Overall, these results validate LLM-as-a-judge as a reliable proxy for human assessment in our setting.


Table~\ref{tab:human_eval_detailed} reports detailed human evaluation results. Vector steering consistently improves dialect authenticity and reduces MSA formality across nearly all settings, with fluency and coherence largely preserved. Neuron steering shows more modest and less consistent gains. Inter-annotator agreement (Table~\ref{tab:human_eval_agreement}) is highest for dialect authenticity ($\kappa_w = 0.760$, 86.1\% within one point) and MSA formality ($\kappa_w = 0.673$, 82.7\% within one point). Coherence and fluency show lower but acceptable agreement, with within-one rates above 79\%, reflecting their inherent subjectivity rather than large systematic disagreements.

\begin{table*}[t]
\centering
\scriptsize
\setlength{\tabcolsep}{3pt}
\begin{tabular*}{\textwidth}{@{\extracolsep{\fill}}lcccc@{}}
\toprule
Metric & $\kappa_w$ & MAD & Exact & Within 1 \\
\midrule
Dialect authenticity & 0.760 & 0.585 & 0.621 & 0.861 \\
Coherence            & 0.512 & 0.837 & 0.482 & 0.792 \\
Arabic fluency       & 0.506 & 0.535 & 0.639 & 0.879 \\
MSA formality        & 0.673 & 0.717 & 0.529 & 0.827 \\
\bottomrule
\end{tabular*}
\caption{
Inter-annotator agreement for human evaluation. $\kappa_w$ denotes
quadratic-weighted Cohen's $\kappa$. MAD is mean absolute difference between
annotator scores. Exact is exact-score agreement, and Within 1 is the fraction
of items where annotators differ by at most one point.
}
\label{tab:human_eval_agreement}
\end{table*}

Overall, the agreement results indicate that both annotators and the LLM judge
are most reliable on the dialect-specific axes. Authenticity shows the strongest
LLM--human agreement, and MSA formality also has high rank correlation, suggesting
that the judge captures the main dialect-control signal. By contrast, fluency
and coherence show weaker LLM--human agreement, likely because these dimensions
are more subjective and less directly tied to the steering intervention. Tables~\ref{tab:qualitative_allam} and~\ref{tab:qualitative_fanar} show representative unsteered, neuron-steered, and vector-steered outputs for ALLaM-7B-Instruct and Fanar-1-9B-Instruct across all four evaluated dialects, illustrating the register shifts underlying the quantitative results in Section~\ref{sec:results}.

\begin{table*}[t]
\centering\small
\setlength{\tabcolsep}{4pt}
\renewcommand{\arraystretch}{1.2}
\begin{tabular}{@{}l p{0.87\textwidth}@{}}
\toprule
\multicolumn{2}{@{}l@{}}{\textbf{Egyptian}} \\
\textit{Prompt} & \ar{عيد صياغة دا بألفاظك بس يكون المعنى الأصلي موجود: "كتير منا بيلاقوا نفسهم وهما بيسمعوا برنامج تلفزيوني بيحكي لنا عن تجربة أو عملية مش هنشاركك فيها أو نطبقها بعد كده."} \\
 & \footnotesize\textit{``Rephrase this in your own words, keeping the original meaning: `Many of us find ourselves listening to a TV programme that tells us about an experience or a procedure we won't take part in or apply afterwards.'\,''} \\
\addlinespace[2pt]
Unsteered & \ar{كثير منا يجد نفسه أثناء الاستماع لبرنامج إذاعي يتحدث عن تجربة أو عملية لن نتمكن من مشاركتها أو تطبيقها لاحقًا.} \\
Neuron & \ar{في كتير مننا بيلاقوا حالهم بيسمعوا برنامج تلفزيوني بيحكولنا عن تجربة أو عملية، من غير ما نكون بنقدر اننا نطبقها أو نشارك فيها بعد كده. يعني، في تجارب أو عمليات، في منها ما بنقدر اننا نطبقها أو نشارك فيها، وبنسمع عنها من غير ما نقدر اننا نطبقها أو نشارك فيها.} \\
Vector & \ar{كتير مننا بيسمعوا عن تجارب أو عمليات من خلال برامج تلفزيونية، بس مش دايما بيقدروا يطبقوها في حياتهم.} \\
\midrule
\multicolumn{2}{@{}l@{}}{\textbf{Moroccan}} \\
\textit{Prompt} & \ar{"ممكن تخصم ليهادشي لعشرين دولارا، عافاك؟" عاود صياغة هدشي بالكلمات دياولك، محافظ على المفهوم الاساسي} \\
 & \footnotesize\textit{``Could you discount this to twenty dollars, please?'' Rephrase it in your own words, keeping the core idea.} \\
\addlinespace[2pt]
Unsteered & \ar{ممكن تنقصلي هادشي ل20 دولار، من فضلك؟} \\
Neuron & \ar{أستسمح، بغيت نقول، ممكن تنقص هادشي ل20 دولار، الله يعطيك الخير؟} \\
Vector & \ar{ممكن تخصم ليا هادشي لعشرين دولار، عافاك؟} \\
\midrule
\multicolumn{2}{@{}l@{}}{\textbf{Saudi}} \\
\textit{Prompt} & \ar{اعد كتابة "وان كان ع الذكرى تطمن باخذ جروحي معايا" عشان تصير مناسبة اكثر لجمهور عام} \\
 & \footnotesize\textit{Rewrite ``And if memory brings reassurance, I'll take my wounds with me'' so it suits a general audience.} \\
\addlinespace[2pt]
Unsteered & \ar{الذكريات قد تكون مريحة لكنها تحمل معها ألم الجروح.} \\
Neuron & \ar{"حتى لو كانت الذكرى تُطمئن، سأحمل جروحي معي"} \\
Vector & \ar{الذكريات ممكن تكون مصدر راحة لكن ما تعوض عن الجروح الحقيقية.} \\
\midrule
\multicolumn{2}{@{}l@{}}{\textbf{Syrian}} \\
\textit{Prompt} & \ar{"ولو ضدي العالم كله عنك لا ما بتخلا" رجاع كتوب هلجملة بطريقة تناسب العوام أكتر} \\
 & \footnotesize\textit{``Even if the whole world were against me, I wouldn't give you up'' --- rewrite this to suit ordinary people better.} \\
\addlinespace[2pt]
Unsteered & \ar{"مهما كان كل العالم ضدي، مستحيل أتخلى عنك."} \\
Neuron & \ar{"ولو العالم كله ضدّي، ما بتخلّى عنك."} \\
Vector & \ar{"لو اجتمع كل البشر ضدك، ما رح أقدر أتخلى عنك."} \\
\bottomrule
\end{tabular}
\caption{Qualitative examples for \textbf{ALLaM-7B-Instruct-preview}.}
\label{tab:qualitative_allam}
\end{table*}

\begin{table*}[h!]
\centering\small
\setlength{\tabcolsep}{4pt}
\renewcommand{\arraystretch}{1.2}
\begin{tabular}{@{}l p{0.87\textwidth}@{}}
\toprule
\multicolumn{2}{@{}l@{}}{\textbf{Egyptian}} \\
\textit{Prompt} & \ar{"انت فين يا حبيبي يا ترى" عايز رد جذاب قايم على الجملة دي.} \\
 & \footnotesize\textit{``Where are you, my love, I wonder'' --- I want an appealing reply based on this line.} \\
\addlinespace[2pt]
Unsteered & \ar{"أهلاً بك! أنا هنا، وأتوق لرؤيتك، فالحب لا يحدّه المسافات."} \\
Neuron & \ar{"في انتظارك بكل حُب ولهفة، يا ملاكي! أينما كنت، فأنت دائمًا في قلبي."} \\
Vector & \ar{"أهلاً يا قلبي، أنا هنا بجنبك طول الوقت، مش فارق إن كنت بعيد أو قريب، حبك دايمًا قريب."} \\
\midrule
\multicolumn{2}{@{}l@{}}{\textbf{Moroccan}\quad\footnotesize\textit{(responses truncated)}} \\
\textit{Prompt} & \ar{قاد واحد الجواب لي فيه هاد الجملة كشيحاجا اساسية: "الرحلات كيكونو رْخص مع بزاف ديال الناس…"} \\
 & \footnotesize\textit{Give an answer built around this line: ``Trips are cheaper with lots of people\dots''} \\
\addlinespace[2pt]
Unsteered & \ar{هذه الجملة تشير إلى أن السفر يكون أرخص عندما يكون مع مجموعة كبيرة من الأشخاص، حيث يمكن تقاسم التكاليف بينهم…} \\
Neuron & \ar{الجملة التي ذكرتها تشير إلى مبدأ اقتصادي شائع حيث يمكن أن تكون الرحلات أو الأنشطة الجماعية أقل تكلفة…} \\
Vector & \ar{هاد الجملة تقول ليك إن الرحلات غادي تكون أرخص إذا كنت مع بزاف ديال الناس، علاش إذا كنت وحدك…} \\
\midrule
\multicolumn{2}{@{}l@{}}{\textbf{Saudi}} \\
\textit{Prompt} & \ar{اعد كتابة "قلبي فداك قلي وش اغلى من القلب" باسلوبك الخاص، لكن مع انك تحافظ على المعنى الأساسي} \\
 & \footnotesize\textit{Rewrite ``My heart is your ransom --- tell me, what is dearer than the heart?'' in your own style, keeping the core meaning.} \\
\addlinespace[2pt]
Unsteered & \ar{"أرواحي كلها لك، أخبرني، ما الذي يمكن أن يكون أغلى من القلب؟"} \\
Neuron & \ar{"أهدي لك قلبي، يا من يستحق أكثر من أي ثروة، فما أغلى القلب عندما يكون مقدمًا لمن نحب!"} \\
Vector & \ar{"قلبك أغلى من كل شيء، فداك به، ما أغلى الحب!"} \\
\midrule
\multicolumn{2}{@{}l@{}}{\textbf{Syrian}\quad\footnotesize\textit{(neuron response truncated)}} \\
\textit{Prompt} & \ar{شو الآثار يلي بتترتب على "كمان، بالمناخات الأدفى، بتنمى البكتيريا بسرعة أكبر وبتعيش لفترة أطول برّا الجسم."، إذا حدا حكالك هيك؟} \\
 & \footnotesize\textit{What follows from ``Also, in warmer climates bacteria grow faster and survive longer outside the body'' --- if someone said that to you?} \\
\addlinespace[2pt]
Unsteered & \ar{إذا سمعت هذا، فإن الآثار المحتملة تشمل زيادة خطر العدوى والأمراض الناجمة عن البكتيريا في البيئات الدافئة، حيث يمكن أن تنمو وتبقى على قيد الحياة خارج الجسم لفترة أطول.} \\
Neuron & \ar{إذا سمعت شخصًا يقول أن "في المناخات الأكثر دفئًا، تنمو البكتيريا بشكل أسرع وتعيش لفترة أطول خارج الجسم"، فهذا صحيح إلى حد كبير…} \\
Vector & \ar{إذا حكالك هيك، يعني إنو المناخات الأدفى بتساعد على نمو البكتيريا بسرعة أكتر وبتعيش لفترة أطول برّا الجسم.} \\
\bottomrule
\end{tabular}
\caption{Qualitative examples for \textbf{Fanar-1-9B-Instruct}.}
\label{tab:qualitative_fanar}
\end{table*}

\end{document}